\documentclass[twoside]{article}

\usepackage[accepted]{aistats2020}
\usepackage{amsfonts}
\usepackage[ruled,vlined]{algorithm2e}
\usepackage{graphicx}
\usepackage{subcaption}
\usepackage{xcolor}
\usepackage{tabularx,booktabs}
\usepackage{textcomp}
\usepackage{multirow}
\usepackage{setspace}
\usepackage{xcolor}
\definecolor{orange}{rgb}{0.0,0.0,0.75}

\makeatother
\usepackage[colorlinks=true]{hyperref}
\hypersetup{
	pdfborder={0 0 0.4},      
	urlbordercolor=red,
	citecolor=orange,
}
\let\oldnl\nl
\newcommand{\nonl}{\renewcommand{\nl}{\let\nl\oldnl}}
\usepackage[separate-uncertainty=true, multi-part-units=repeat]{siunitx}
\newcolumntype{C}{>{\centering\arraybackslash}X} 
\graphicspath{{./Figures/}}
\definecolor{edited}{rgb}{0.0, 0.0, 0.0}
\definecolor{added}{rgb}{0.0, 0.0, 0.0}
\definecolor{fe}{rgb}{0.0, 0.0, 0.0}

\usepackage[round]{natbib}

\begin{document}

%

%

\twocolumn[

\aistatstitle{Invertible Generative Modeling using Linear Rational Splines}

\aistatsauthor{Hadi M. Dolatabadi \And Sarah Erfani \And Christopher Leckie}
\vspace*{0.33em}
\aistatsaddress{School of Computing and Information Systems\\ The University of Melbourne} ]

\begin{abstract}
	
Normalizing flows attempt to model an arbitrary probability distribution through a set of invertible mappings. These transformations are required to \textcolor{edited}{achieve} a tractable Jacobian determinant \textcolor{edited}{that can be used} in high-dimensional scenarios. \textcolor{edited}{The first normalizing flow designs used} coupling layer mappings built upon affine transformations. The significant advantage of such models is their easy-to-compute inverse. Nevertheless, making use of affine transformations \textcolor{edited}{may} limit the expressiveness of such models. \textcolor{edited}{Recently,} invertible piecewise polynomial functions as a replacement for affine transformations have attracted attention. However, these methods require solving a polynomial equation to calculate their inverse. In this paper, we explore using linear rational splines as a replacement for affine transformations used in coupling layers. Besides having a straightforward inverse, inference and generation have similar cost and architecture in this method. Moreover, simulation results \textcolor{edited}{demonstrate} the competitiveness of this approach's performance \textcolor{edited}{compared to} existing methods.   

\end{abstract}

\section{INTRODUCTION}

Flow-based modeling, widely known as normalizing flows~\citep{tabak2013family, rezende2015variational}, is a novel approach \textcolor{fe}{used for density estimation} problems. The main idea behind this method is to model the distribution of any arbitrary set of data as the mapping of a simple base random variable using a set of invertible transformations. In doing so, they make use of the well-known change of variables formula from probability theory~\citep{papoulis2002probability}. In particular, let $\mathbf{Z}$ denote a random vector with a simple distribution such as a standard normal. Furthermore, \textcolor{edited}{let} $\mathbf{X}$ \textcolor{edited}{be} the result of applying an invertible transformation $\mathbf{f}(\cdot)$ on $\mathbf{Z}$. Then, it can be shown that:
\begin{equation}\label{eq:chng_vrbl}
	p_{\mathbf{X}}(\mathbf{x})=p_{\mathbf{Z}}(\mathbf{z})\left|\mathrm{det}\Big(\dfrac{\partial \mathbf{f}}{\partial \mathbf{z}}\Big)\right|^{-1},
\end{equation}
where $p_{\mathbf{X}}(\mathbf{x})$ and $p_{\mathbf{Z}}(\mathbf{z})$ denote the probability distributions of $\mathbf{X}$ and $\mathbf{Z}$, respectively. Here, $\left|\mathrm{det}\Big(\dfrac{\partial \mathbf{f}}{\partial \mathbf{z}}\Big)\right|$ indicates the absolute value of the \textit{Jacobian determinant} for the transformation $\mathbf{f}(\cdot)$. This quantity captures the amount of change in $\mathbf{z}$ caused by the transformation (\textit{flow}) $\mathbf{f}(\cdot)$ through the Jacobian operation. Then, this change in variable $\mathbf{z}$ \textcolor{edited}{can be} used to \textit{normalize} the probability distribution of the data through \textcolor{edited}{the} determinant calculation.

Eq. \eqref{eq:chng_vrbl} can be used for maximum likelihood problems. \textcolor{edited}{This can be done by} parameterizing the distribution of the data through an invertible mapping $\mathbf{f}(\cdot)$ and \textcolor{edited}{then} finding the parameters of this transformation using an \textcolor{edited}{appropriate} optimization algorithm. 

However, a major bottleneck to \textcolor{edited}{using} normalizing flows in high-dimensional scenarios is the complexity of \textcolor{edited}{computing} the Jacobian determinant. In general, determinant calculation has an \textcolor{fe}{$\mathcal{O}\big(D^3\big)$} complexity for an arbitrary \textcolor{fe}{$D\times D$} matrix. Hence, to adapt flow-based models to high-dimensional cases, this issue must be addressed. As a result, the main objective in designing normalizing flow algorithms is to come up with an invertible transformation whose Jacobian determinant is tractable. 

This work presents a new transformation for use in normalizing flow algorithms. We \textcolor{edited}{show} that the transformation used has an analytical inverse. Moreover, experiments done using this transformation indicate its competitive performance with existing state-of-the-art algorithms. 


\section{RELATED WORK}\label{sec:bkg_lr}

In this section, we review the existing methods for flow-based modeling.

\subsection{Coupling Layer Transformations}

The main idea behind this category of transformations is as follows. They first split the data into two \textcolor{edited}{parts}. The first part \textcolor{edited}{is} output without any change. Then, a transformation is \textcolor{edited}{constructed} based on the first partition of the data. This \textcolor{edited}{transformation} is then applied to the second \textcolor{edited}{part} of the data \textcolor{edited}{to generate} the output.

Specifically, \textcolor{edited}{let $\mathbf{z} \in \mathbb{R}^{\mathrm{D}}$ show the data to be transformed}. First, it \textcolor{edited}{is} split into two parts $\mathbf{z}_1 \in \mathbb{R}^{\mathrm{d}}$ and $\mathbf{z}_2\in \mathbb{R}^{\mathrm{D-d}}$ where $d<D$. Then, each part is transformed separately using
\begin{align}\label{eq:cplng_lyr}
\mathbf{x}_1 &= \mathbf{z}_1 \nonumber \\
\mathbf{x}_2 &= \mathbf{g}_{\boldsymbol{\theta}({\mathbf{x}_1})}(\mathbf{z}_2),
\end{align}
where $\mathbf{g}_{\boldsymbol{\theta}({\mathbf{x}_1})}(\cdot)$ is an invertible, \textcolor{fe}{element-wise} transformation with parameters $\boldsymbol{\theta}({\mathbf{x}_1})$ which \textcolor{edited}{have} been computed based on $\mathbf{x}_1$. The final output of the transformation is \textcolor{edited}{then} $\mathbf{x}=[\mathbf{x}_1, \mathbf{x}_2]$. It can be shown that such transformations have a lower triangular Jacobian whose determinant is the multiplication of the diagonal elements. Note that coupling layers do not change the first split of their inputs. Thus, to prevent that part of the data from remaining unchanged, it is necessary to use an alternating mask and switch the splits at two consecutive transformations to make sure that every dimension of the data \textcolor{edited}{has} a chance to be changed.

\textcolor{edited}{The} NICE algorithm~\citep{dinh2015nice} uses a simple translation function as its transformation $\mathbf{g}_{\boldsymbol{\theta}({\mathbf{x}_1})}(\cdot)$. \textcolor{edited}{It sets} $\mathbf{g}_{\boldsymbol{\theta}({\mathbf{x}_1})}(\mathbf{z}_2) = \mathbf{z}_2 + \boldsymbol{\theta}({\mathbf{x}_1})$ where $\boldsymbol{\theta}(\cdot)$ is a multi-layer perceptron (MLP) with Rectified Linear Units (ReLU) as the activation function.

Real~NVP~\citep{dinh2016density} generalizes NICE by using an affine transformation as $\mathbf{g}_{\boldsymbol{\theta}({\mathbf{x}_1})}(\cdot)$. Here, the authors use two Residual Networks (ResNet)~\citep{he2016deep} to come up with the translation and scaling operations required for an affine mapping. As in NICE, Real~NVP uses a simple alternating mask \textcolor{edited}{to involve} every dimension of the data in the transformation. In \cite{kingma2018glow}, it is suggested to \textcolor{fe}{linearly combine dimensions} of the data before feeding it to each layer of transformation using an invertible \textcolor{fe}{$1 \times 1$} convolution operation, resulting in a new algorithm called Glow. This permutation is simply a matrix multiplication. Thus, to make \textcolor{edited}{the} Glow algorithm fast, the authors propose using an LU-decomposition to calculate the matrix mentioned above. It is shown that Glow~\citep{kingma2018glow} \textcolor{fe}{improves} {Real~NVP's} \textcolor{fe}{performance}~\citep{dinh2016density} for \textcolor{fe}{generative image modeling}.

Despite its better performance compared to Real~NVP, Glow \textcolor{fe}{may struggle to learn} synthetic probability distributions with multiple separated modes~\citep{grathwohl2019FFJORD}. Since all the previously mentioned algorithms use an affine transformation, this \textcolor{edited}{may} be the cause of the degradation in their \textcolor{edited}{general expressiveness}.  

\subsection{Autoregressive Transformations}

The chain rule in probability theory states that if ${\mathbf{Z}=\big(Z_1, Z_2, \dots, Z_D\big)}$ is a $D$-dimensional random variable, then the distribution of ${\mathbf{Z}}$ can be written as~\citep{papoulis2002probability}
\begin{equation}\label{eq:chn_rl}
	p_{\mathbf{Z}}(\mathbf{z})=p_{Z_1}(z_1)\prod_{i=2}^{D}p_{Z_i|\mathbf{Z}_{<i}}(z_i|\mathbf{z}_{<i}),
\end{equation}
where $\mathbf{z}_{<i}$ is shorthand notation denoting all dimensions of vector $\mathbf{z}$ with an index less than $i$.

Autoregressive flows exploit this rule to build their transformations. In particular, they transform each one-dimensional variable $z_i$ conditioned on its previous dimensions $\mathbf{z}_{<i}$ using an invertible transformation. It can be shown that the Jacobian determinant of such transformations is again lower triangular and can be computed efficiently.

Inverse Autoregressive Flows (IAF)~\citep{kingma2016improving} and Masked Autoregressive Flows (MAF)~\citep{papamakarios2017masked} are among the first designs in this category. They use affine functions to build their autoregressive transformation. Later, it was argued that the use of affine functions might limit the expressiveness of such models~\citep{huang2018neural}. Hence, Neural Autoregressive Flows (NAF)~\citep{huang2018neural} were introduced. NAFs build their transformation using a neural network with positive weights \textcolor{fe}{and monotonic activations} to ensure invertibility. A hyper neural network, \textcolor{added}{called the conditioner network,} is trained to capture the dependency of the neural network parameters to the previously seen data. Later, Block Neural Autoregressive Flows (B-NAF)~\citep{de2019block} suggest a more straightforward structure omitting the conditioner network. It was shown that B-NAFs reach the same performance as NAFs using orders of magnitude fewer parameters. Moreover, universal density approximation \textcolor{edited}{has been} proved for both NAFs and B-NAFs. This theorem states that a structure of these models always exists that could accurately represent the data.

Despite their success, there are two important observations regarding NAFs and B-NAFs. \textit{First, universal density approximation only proves the existence of such models without any convergence guarantee.} Moreover, these two methods are not analytically invertible, questioning their usage for generative probabilistic modeling tasks where one might want to compute the inverse. 

In general, unlike coupling layer transformations, inverting an autoregressive flow cannot be done in a single pass, making their inverse computationally expensive. Thus, it would be better if one can come up with a coupling layer transformation whose performance can compete with autoregressive flows. There are also other autoregressive models \textcolor{edited}{to which} we refer the interested reader~\citep{germain2015made, chen2017var, oliva2018tan}.

\subsection{Other Methods}

\textcolor{edited}{In addition to} the coupling layer and autoregressive transformations, there are other methods that do not fall into any of the previous categories. We \textcolor{edited}{summarize} the most well-known ones here.

Continuous Normalizing Flows (CNF) are flow-based models that are constructed upon Neural Ordinary Differential Equations (ODE)~\citep{chen2018node}. FFJORD~\citep{grathwohl2019FFJORD} improves CNF's computational complexity using the Hutchinson's trace estimator~\citep{hutchinson1990stochastic}. Both of these methods involve a system of first-order ODEs that replaces the change of variables formula in Eq. \eqref{eq:chng_vrbl}.  Then, this ODE system is solved by a proper integrator resulting in a flow-based model. Here, the sampling process requires solving a system of ODEs, which may slow down \textcolor{fe}{sample generation}.

A closely related method to FFJORD is invertible residual networks, or i-ResNets for short~\citep{behrmann2019iresnet}. In this paper, the authors first state \textcolor{fe}{a sufficient condition for making} residual networks invertible. Then, a flow-based generative model is built using this invertible transformation. Realization of such models depends on calculating the Jacobian determinant of an entire residual network. In \cite{behrmann2019iresnet}, it is shown that an infinite power series can replace this Jacobian determinant. To compute this infinite series feasibly, the authors suggest truncating it after a finite number of terms $n$, which causes the Jacobian determinant estimator to become biased~\citep{behrmann2019iresnet}.

Residual Flows~\citep{chen2019residualflows} address the bias issue of i-ResNets by using a so-called \textit{Russian roulette} estimator~\citep{kahn1955use}. In short, instead of a deterministic $n$ \textcolor{added}{used for truncation of the infinite series} as in i-ResNets, a \textit{Russian roulette} estimator models $n$ as a discrete random variable on natural numbers. Then, the infinite power series is replaced with the first $n$ terms divided by \textcolor{fe}{appropriate weights}. Unlike i-ResNets, here $n$ is a realization of an arbitrary distribution on natural numbers. Since Residual Flows use an unbiased estimator for the Jacobian determinant, it is shown that they can achieve a better performance compared to i-ResNets~\citep{behrmann2019iresnet}.

A major drawback of models such as i-ResNets and Residual Flows, which make use of invertible residual networks as their building blocks, is their inversion. Although i-ResNets are guaranteed to be invertible, they do not have an analytical inverse. I-ResNets, as well as Residual Flows, need a fixed-point iterative algorithm to compute the inverse of each layer. Hence, the inversion process takes much more time (around 5-20x) than inference~\citep{behrmann2019iresnet}.

\textcolor{edited}{\section{PROPOSED APPROACH}\label{sec:lrs_vs_rq}}

\subsection{Neural Spline Flows}
As we saw in Eq. \eqref{eq:cplng_lyr}, the only requirements that $\mathbf{g}_{\boldsymbol{\theta}({\mathbf{x}_1})}(\cdot)$ must satisfy is being invertible and differentiable. Also, we saw that Real~NVP and Glow use \textcolor{fe}{a simple affine transformation}, to build their algorithm and maintain analytical invertibility. \textcolor{edited}{In contrast}, autoregressive methods either use a simple affine transformation like IAFs and MAFs, or use non-analytically invertible neural networks with positive weights as in NAFs and B-NAFs. However, there are a plethora of differentiable functions that are invertible \textcolor{edited}{and lie} in-between: they can be more expressive than a simple affine mapping while having an easy-to-compute inverse.

A family of functions with such properties is splines. Splines are piecewise functions where each piece is expressed as a closed-form standard function. The most popular form of splines is those defined by polynomials.

In the context of flow-based modeling, the idea of replacing the affine transformation used in methods such as NICE with piecewise polynomial functions was first introduced by \cite{mueller2019neural}. They use a piecewise linear or quadratic function as a replacement for the affine transformation in coupling layers. It is shown in that this change \textcolor{edited}{improves} the expressiveness of methods such as Real~NVP.

Later, this work was extended to the cases of cubic and rational quadratic splines~\citep{durkan2019cubic, durkan2019neural}. It is shown that the previous coupling layer methods such as Real~NVP and Glow could benefit from this change. Also, simulation results demonstrate their competitive performance against the most expressive methods, such as NAF and B-NAF, without sacrificing analytical invertibility.

In this work, we explore the usage of another family of piecewise functions, namely linear rational splines, in the context of normalizing flows. We aim to seek more straightforward piecewise functions whose inversion \textcolor{edited}{can be} done efficiently. Previous methods in this area use quadratic, cubic, and rational quadratic functions whose inversion is done after solving degree 2 or 3 polynomial equations. However, piecewise linear rational splines can perform competitively with these methods without requiring a polynomial equation to be solved in the inversion.

\textcolor{added}{Next, we review linear rational splines and their application in constructing monotonically increasing functions. Then, based on this algorithm, we propose our flow-based model.} 

\subsection{Monotonic Data Interpolation using Linear Rational Splines}

\textcolor{edited}{Let} $\big\{\big(x^{(k)},~y^{(k)}\big)\big\}_{k=0}^{K}$ be a set of monotonically increasing \textcolor{fe}{points} called knots. Furthermore, let ${\big\{d^{(k)}>0\big\}_{k=0}^{K}}$ be a set of positive numbers representing the derivative of each point. Consider that we wish to find linear rational functions\footnote{In the context of splines, they are also known as linear/linear rational functions. Also, they are sometimes referred to as homographic functions.} of the form $y=\tfrac{ax+b}{cx+d}$ that fit the given points and their respective derivatives in each interval (also called a bin) $\big[x^{(k)}, x^{(k+1)}\big]$. Also, \textcolor{edited}{consider that we require} the function to be monotone.

In each bin, after satisfying function value constraints at the start and end points, we can write down the desired function as:
\begin{equation}\label{eq:lrs}
g(x)=\dfrac{w^{(k)} y^{(k)} (1-\phi) + w^{(k+1)} y^{(k+1)} \phi}{w^{(k)} (1-\phi) + w^{(k+1)} \phi},
\end{equation}
in which $w^{(k)}$ and $w^{(k+1)}$ are two arbitrary weights and $\phi=\big(x-x^{(k)}\big)/\big(x^{(k+1)}-x^{(k)}\big)$, which belongs to the interval $[0,1]$. As can be seen in Eq. \eqref{eq:lrs}, we only have one degree of freedom left while \textcolor{edited}{still needing} to satisfy two derivative constraints at the extreme points of the bin. \textcolor{added}{Thus, we cannot use a single linear rational function and satisfy all the constraints of each bin}. 

\cite{fuhr1992monotone} suggest to solve this issue by considering an intermediate point in each interval $\big(x^{(k)}, x^{(k+1)}\big)$. This way, we \textcolor{edited}{can} treat each bin as it was two. Since there are no constraints on this particular intermediate point in terms of \textcolor{edited}{the} value and derivative, we can use its associated parameters to add more degrees of freedom to the existing ones. Then, we can fit one linear rational function to each one of these two intervals and satisfy the end point values and derivative constraints.

In particular, let $x^{(m)} = (1-\lambda) x^{(k)} + \lambda x^{(k+1)}$. When $0<\lambda<1$, this equation denotes a point in the interval $\big(x^{(k)}, x^{(k+1)}\big)$. As in \cite{fuhr1992monotone}, we \textcolor{edited}{aim} to fit two linear rational functions like Eq. \eqref{eq:lrs} to the intervals $\big[x^{(k)}, x^{(m)}\big]$ and $\big[x^{(m)}, x^{(k+1)}\big]$. Here, the value, weight and derivative of the intermediate point are treated as parameters to satisfy the value and derivative constraints that we have at each bin.

Our desired function is required to be continuous and monotonic. Since \textcolor{edited}{the} derivative of a linear rational function does not change its sign, when we interpolate it using positive derivatives, this constraint is always satisfied. Hence, we only need to take care of the continuity of the function and its derivative at $x^{(m)}$.

After satisfying all those constraints, the final algorithm for monotonic data interpolation using linear rational functions for each interval $\big[x^{(k)}, x^{(k+1)}\big]$ can be shown as in Algorithm \ref{Algorithm}~\citep{fuhr1992monotone}. We can then use this algorithm for all the bins, and end up having a piecewise monotonically increasing function, also known as a linear rational spline. \textcolor{edited}{Specifically}, for each bin $\big[x^{(k)}, x^{(k+1)}\big]$ we have:
\begin{equation}\label{eq:lrs_final}
\scalebox{0.8}{$g(\phi)=\begin{cases} 
	\hfil \dfrac{w^{(k)} y^{(k)} \big(\lambda^{(k)}-\phi\big) + w^{(m)} y^{(m)} \phi}{w^{(k)} \big(\lambda^{(k)}-\phi\big) + w^{(m)} \phi} & 0 \leq \phi \leq \lambda^{(k)} \\ \\
	\dfrac{w^{(m)} y^{(m)} \big(1-\phi\big) + w^{(k+1)} y^{(k+1)} \big(\phi - \lambda^{(k)}\big)}{w^{(m)} \big(1-\phi\big) + w^{(k+1)} \big(\phi-\lambda^{(k)}\big)} & \lambda^{(k)} \leq \phi \leq 1
	\end{cases}$}
\end{equation}

\begin{algorithm*}[tb!]\setstretch{1.35}
	
	\KwIn{$x^{(k)}<x^{(k+1)}$, $y^{(k)}<y^{(k+1)}$, $d^{(k)}>0$ and $d^{(k+1)}>0$}
	\KwOut{$\lambda^{(k)}$, $w^{(k)}$, $w^{(m)}$, $w^{(k+1)}$ and $y^{(m)}$ (Parameters of two homographic functions of Eq. \ref{eq:lrs})}
	
	\nonl Set $w^{(k)}>0$ and $0<\lambda^{(k)}<1$\;
	
	\nonl Calculate $w^{(k+1)}=\sqrt{{d^{(k)}}/{d^{(k+1)}}} w^{(k)}$\;
	
	\nonl Compute $y^{(m)}=\dfrac{\big(1-\lambda^{(k)}\big)w^{(k)}y^{(k)} + \lambda^{(k)} w^{(k+1)}y^{(k+1)}}{\big(1-\lambda^{(k)}\big)w^{(k)} + \textcolor{fe}{\lambda^{(k)}} w^{(k+1)}}$\;
	
	\nonl Calculate $w^{(m)}=\Big(\lambda^{(k)} w^{(k)} d^{(k)} + \big(1-\lambda^{(k)}\big) w^{(k+1)} d^{(k+1)}\Big)\dfrac{x^{(k+1)} - x^{(k)}}{y^{(k+1)} - y^{(k)}}$ 
	
	\caption{{\bf~\citep{fuhr1992monotone} Linear Rational Spline Interpolation of Monotonic Data in the Interval $\big[x^{(k)}, x^{(k+1)}\big]$} \label{Algorithm}}
\end{algorithm*}

\begin{figure}[tb!]
	\centering
	\includegraphics[width=0.4\textwidth]{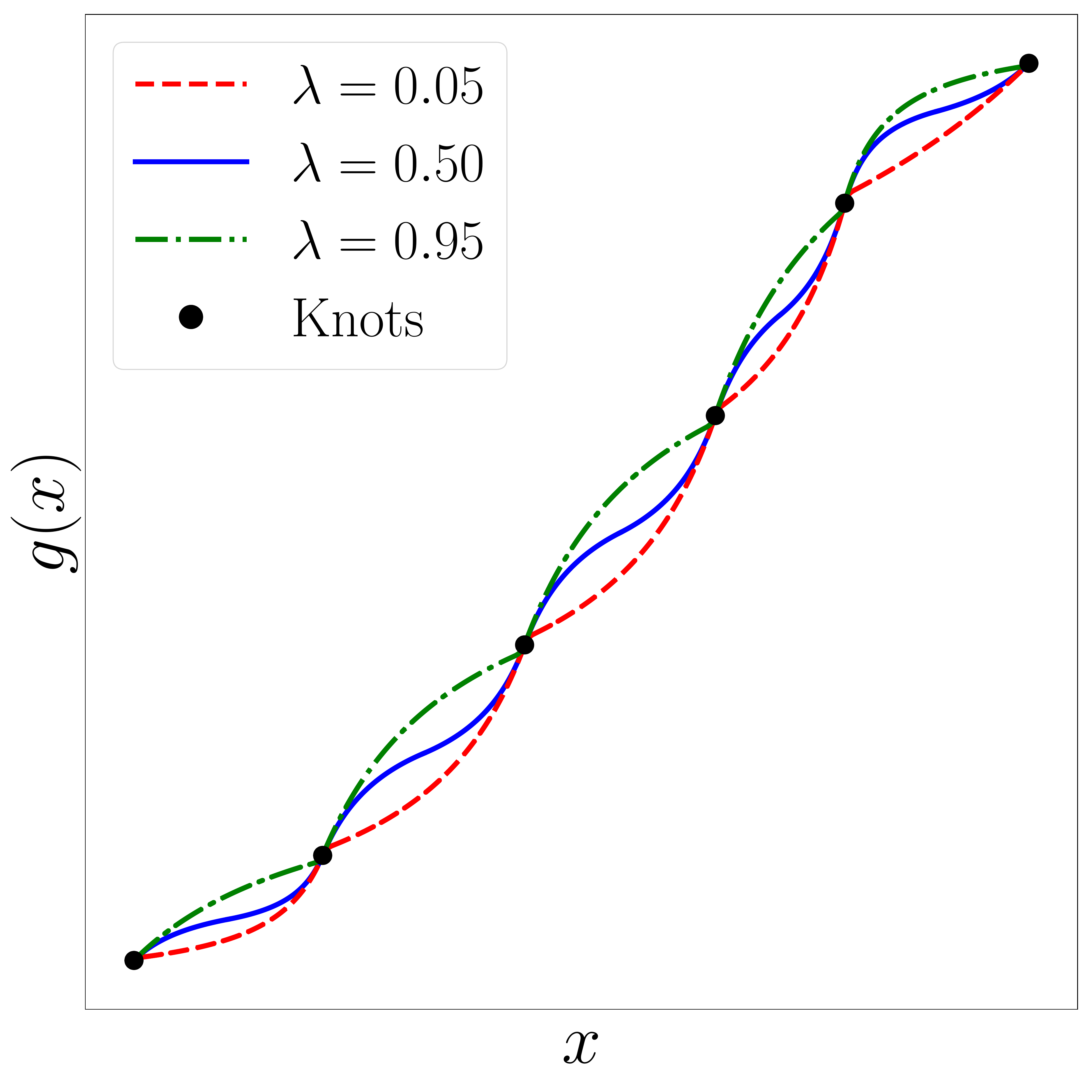}
	\caption{Monotonic data interpolation using linear rational splines. As it can be seen, different settings of $\lambda$ \textcolor{edited}{can} result in different curves.}
	\label{fig:lambda}
\end{figure}

\textcolor{added}{By having a monotonically increasing function whose derivatives exist at all points, we can use it as an alternative for the function used in Eq. \eqref{eq:cplng_lyr} and construct a flow-based model}. Unlike previously used piecewise functions, linear rational splines have the advantage \textcolor{edited}{of having a straightforward inverse that \textcolor{fe}{does} not require solving} a degree 2 or 3 polynomial equation. Moreover, \textcolor{added}{the inverse of linear rational functions} has the same format as its forward \textcolor{edited}{form}, but with different parameters. Thus, the regular and inverse function evaluations cost the same. More interestingly, having one extra degree of freedom (namely $\lambda^{(k)}$\footnote{Note that although $w^{(k)}$ is chosen freely in Algorithm \ref{Algorithm}, since $w^{(m)}$ and $w^{(k+1)}$ are a multiplication of $w^{(k)}$, this \textcolor{edited}{value} does not provide any degree of freedom.}) per bin provides the opportunity to manipulate the curve that fits through a set of possible knots and derivatives. This flexibility is shown in Figure \ref{fig:lambda}. Here, a set of knot points with fixed derivatives are interpolated with different $\lambda^{(k)}$s. In this figure, the assumption is that a single function shares the same \textcolor{edited}{value} for all $\lambda^{(k)}$s. However, \textcolor{edited}{this does not need to be the case}, and one could manipulate $\lambda^{(k)}$ of each bin separately, resulting in \textcolor{edited}{greater} flexibility.

\textcolor{added}{Note that spline transformations are defined within a finite interval. To deal with unbounded data, there are two possible solutions. First, a transformation can be used to map the unbounded data to the desired range. For instance, \cite{mueller2019neural} and \cite{durkan2019cubic} suggest mapping the data into a $[0, 1]$ interval to deal with this issue. A downside to this method is the numerical errors caused by this extra transformation~\citep{durkan2019cubic}. As an alternative approach, \cite{durkan2019neural} propose to use linear tails outside of the interval defined by piecewise functions. This way, the transformation range covers all the real line, circumventing the need to force the data itself to be in the interval $[0, 1]$. We also prefer this way to deal with this problem}.

\subsection{Linear Rational Spline (LRS) Flows}

Having an algorithm for monotone data interpolation, \textcolor{edited}{we can} exploit and adapt it to coupling layers, \textcolor{edited}{and thus}, come up with a normalizing flow. We follow the steps of neural spline flows~\citep{durkan2019neural}.

 First, $K$ and $B$, the number of bins and the boundary of the spline calculation are set. Then, in order to compute the transformation as in Eq. \eqref{eq:cplng_lyr}, a neural network such as a ResNet should be trained to determine the parameters of the transformation $\mathbf{g}_{\boldsymbol{\theta}({\mathbf{x}_1})}(\cdot)$. As we have seen in Algorithm \ref{Algorithm}, $4K-1$ parameters are required for each dimension of this transformation: a width, height, and $\lambda$ for each bin; and $K-1$ derivatives at all points except the start and end points. \textcolor{added}{At these two points} the derivative is set to be $1$ for consistency with linear tails. After determining the transformation parameters for each dimension of the data, we apply the linear rational spline algorithm to come \textcolor{fe}{up} with a set of functions that later construct different dimensions of $\mathbf{g}_{\boldsymbol{\theta}({\mathbf{x}_1})}(\cdot)$. As in Glow, a \textcolor{fe}{$1 \times 1$} convolution constructed by LU-decomposition is also used to \textcolor{fe}{linearly combine} the data that is going to be fed into the coupling layer. \textcolor{fe}{Furthermore, a multi-scale architecture is used for image generation scenarios as in Real~NVP and Glow}.

Note that all spline transformations can also be used in an autoregressive fashion. However, coupling layers are preferred as they can perform the inversion in a single pass. 

Moreover, as \cite{durkan2019neural} suggest, we perform a transformation on the first partition of the data using trainable parameters. \textcolor{fe}{In particular, let $\boldsymbol{\phi}$ be a set of trainable parameters that does not depend on the data. Then, Eq. \eqref{eq:cplng_lyr} can be re-written as
\begin{align}\label{eq:cplng_lyr_new}
\mathbf{x}_1 &= \mathbf{g}_{\boldsymbol{\phi}}(\mathbf{z}_1) \nonumber \\
\mathbf{x}_2 &= \mathbf{g}_{\boldsymbol{\theta}({\mathbf{x}_1})}(\mathbf{z}_2),
\end{align}
where $\mathbf{g}_{\boldsymbol{\psi}}(\cdot)$ is an invertible, element-wise linear rational spline with parameters $\boldsymbol{\psi}$.} This way \textcolor{fe}{all the variables are transformed} while the Jacobian determinant still remains lower triangular.

\textcolor{added}{Before considering the simulation results, it is worthwhile to highlight the differences between this work and previous approaches. First, here the inverse has a straightforward relationship and does not require solving degree 2 or 3 polynomial equations. Second, as in Real~NVP and Glow, the inverse of this transformation is given with a similar format to its forward form: both of them are linear rational splines. This property can be useful in \textcolor{fe}{theoretical analysis of this flow-based model}. For instance, it is sufficient to investigate \textcolor{fe}{mathematical properties (such as the bi-Lipschitz property that can be used for stability guarantees as in i-ResNets~\citep{behrmann2019iresnet})} for a rational linear spline. Then, this property can hold for both the forward and the inverse transformations as both of them are linear rational splines}. \textcolor{fe}{In contrast, rational quadratic splines need slightly fewer parameters. This issue can be avoided by fixing $\lambda$ in linear rational splines.}
    
 \section{SIMULATION RESULTS}\label{sec:sim_res}
 In this section, we review our simulation results. We see that the proposed method can perform competitively despite using a lower order polynomial. \textcolor{fe}{The code is available online at: \href{https://github.com/hmdolatabadi/LRS_NF}{https://github.com/hmdolatabadi/LRS\_NF}}.
 
 \subsection{Synthetic Density Estimation}\label{sec:synth_den}
 
 As a first experiment, we studied a density estimation scenario on synthetic 2-d distributions. This task involves the reconstruction of a continuous probability distribution given a set of its samples. 

\begin{figure*}[tb!]
 \centering
 	\begin{subfigure}{.24\textwidth}
 		\centering
 		\includegraphics[width=1.0\linewidth]{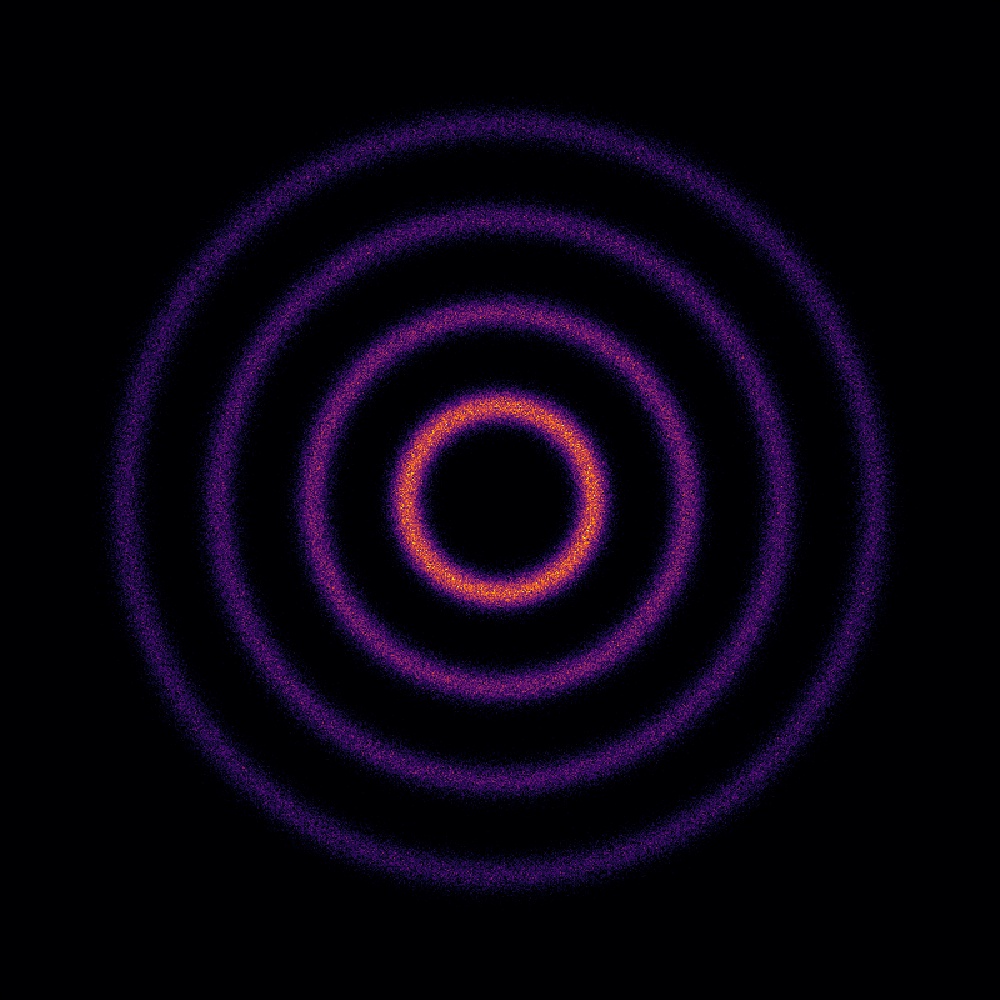}
 		\caption*{Data Samples}
 		\label{fig:density_samples}
 	\end{subfigure}\hspace*{-0.35em}
 	\begin{subfigure}{.24\textwidth}
 		\centering
 		\includegraphics[width=1.0\linewidth]{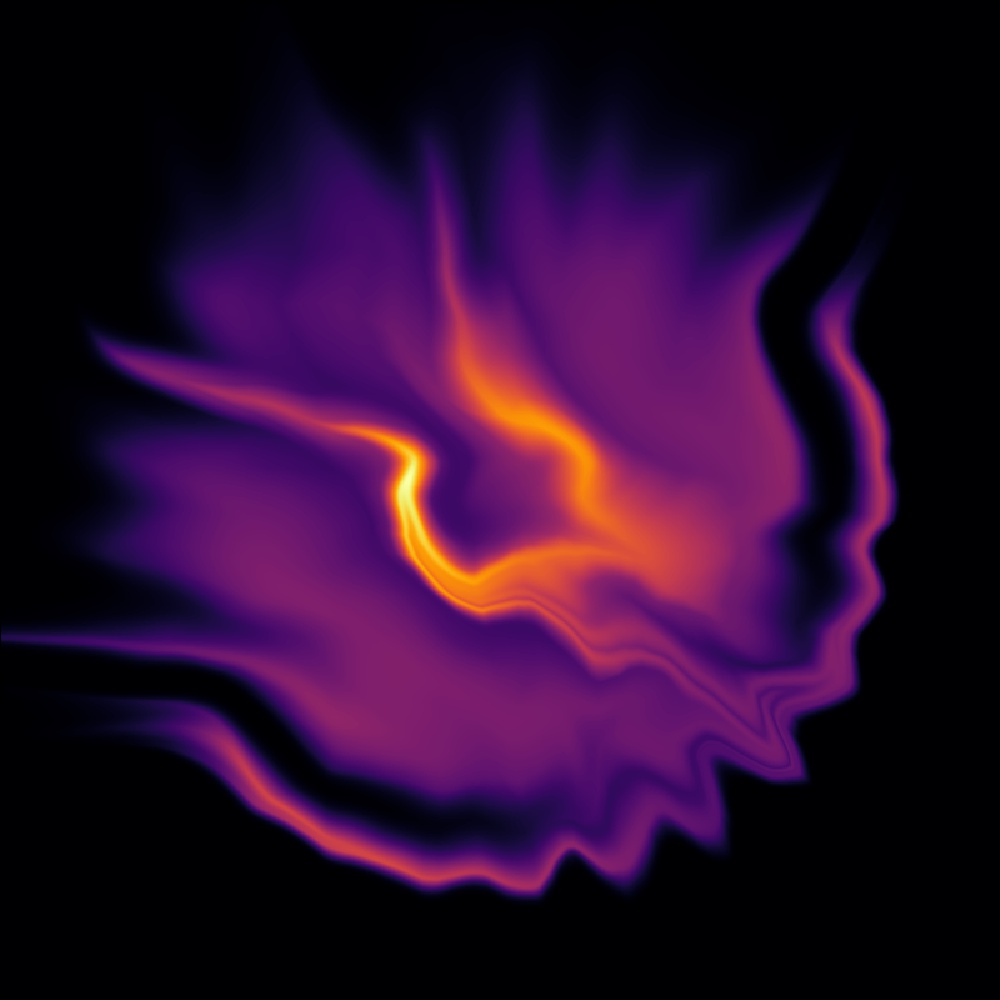}
 		\caption*{Glow}
 		\label{fig:density_glow}
 	\end{subfigure}\hspace*{-0.35em}
 	\begin{subfigure}{.24\textwidth}
 		\centering
 		\includegraphics[width=1.0\linewidth]{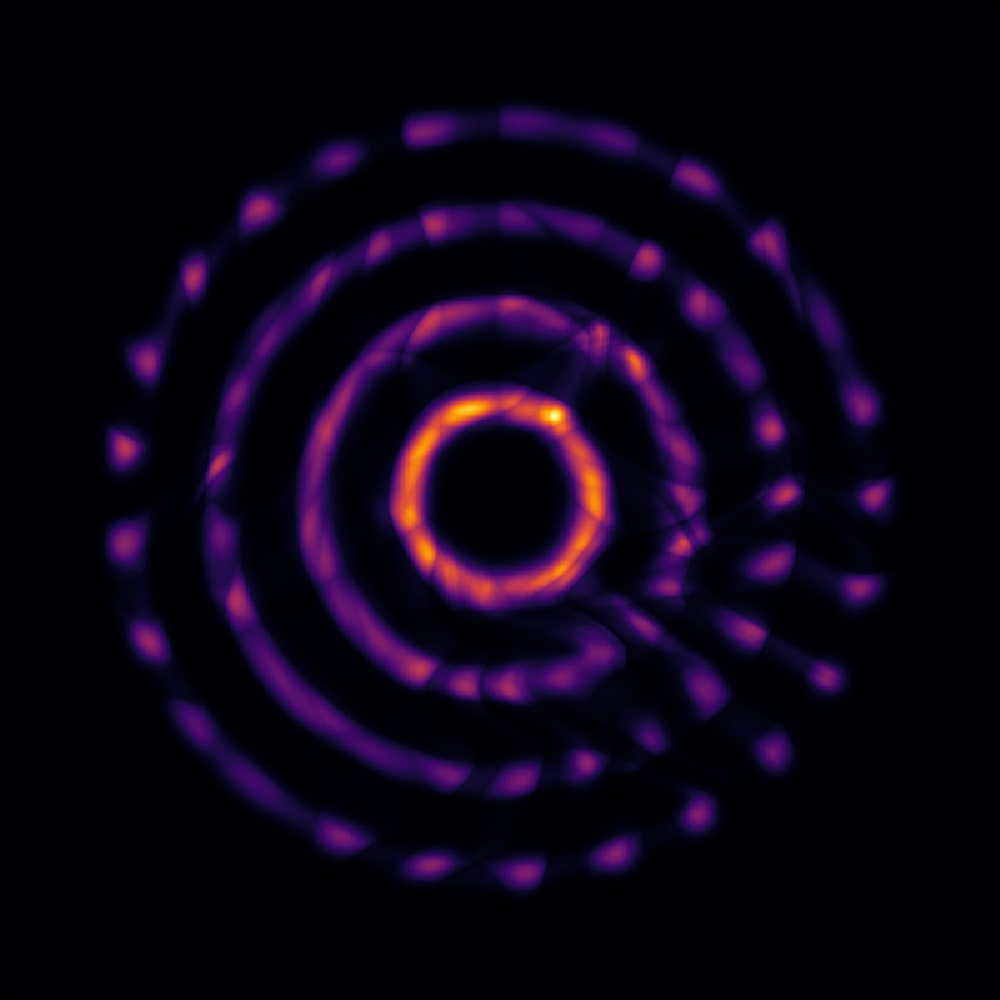}
 		\caption*{i-ResNet}
 		\label{fig:density_iresnet}
 	\end{subfigure}\hspace*{-0.35em}
  	\begin{subfigure}{.24\textwidth}
	 	\centering
	 	\includegraphics[width=1.0\linewidth]{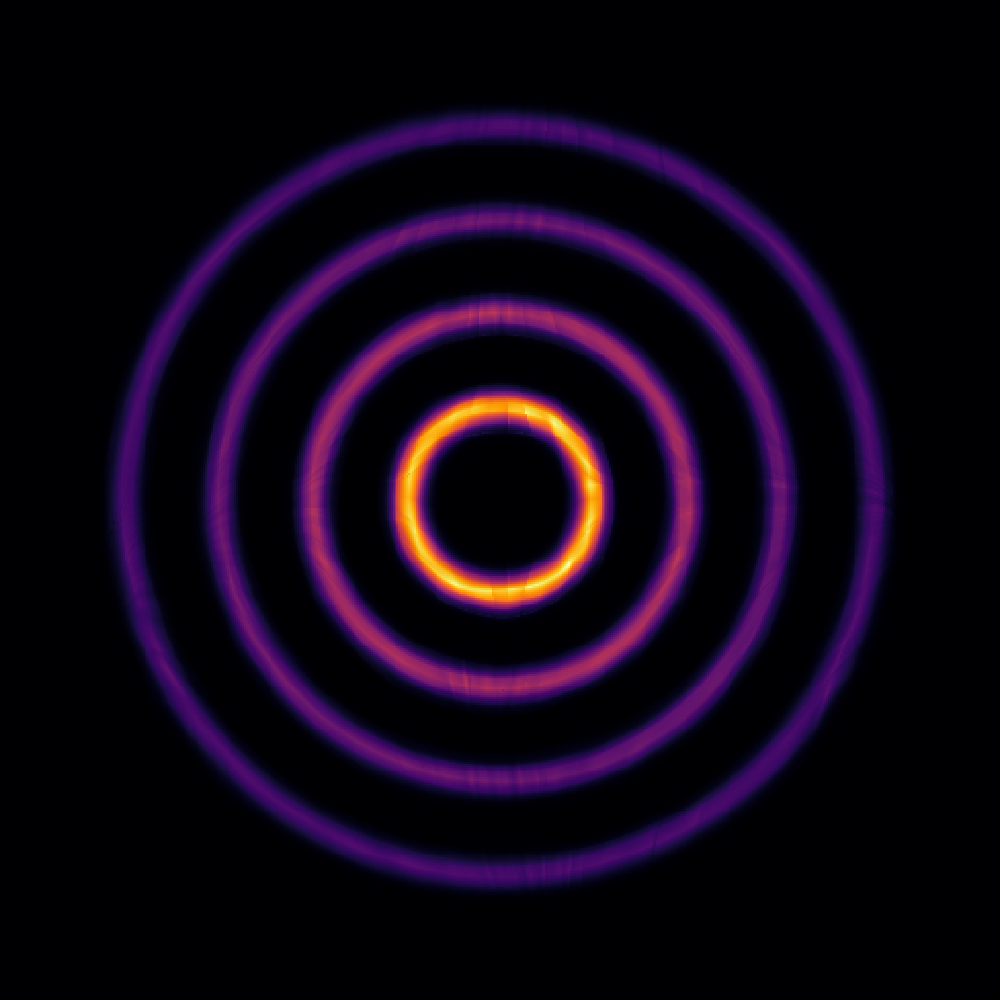}
	 	\caption*{Ours}
	 	\label{fig:density_lrs}
 \end{subfigure}
 	\caption{Density estimation on synthetic 2-d data samples. The first three images were taken from \cite{behrmann2019iresnet}, with permission.}
 	\label{fig:density_estimation}
 \end{figure*}

Figure \ref{fig:density_estimation} compares the performance of Glow~\citep{kingma2018glow}, i-ResNets~\citep{behrmann2019iresnet} and our proposed method under this scenario. Qualitatively, linear rational spline (LRS) flows can reconstruct the underlying distribution precisely, and outperform the other two models. 

\begin{figure*}[tb!]
 \centering
	\begin{subfigure}{.24\textwidth}
		\centering
		\includegraphics[width=1.0\linewidth]{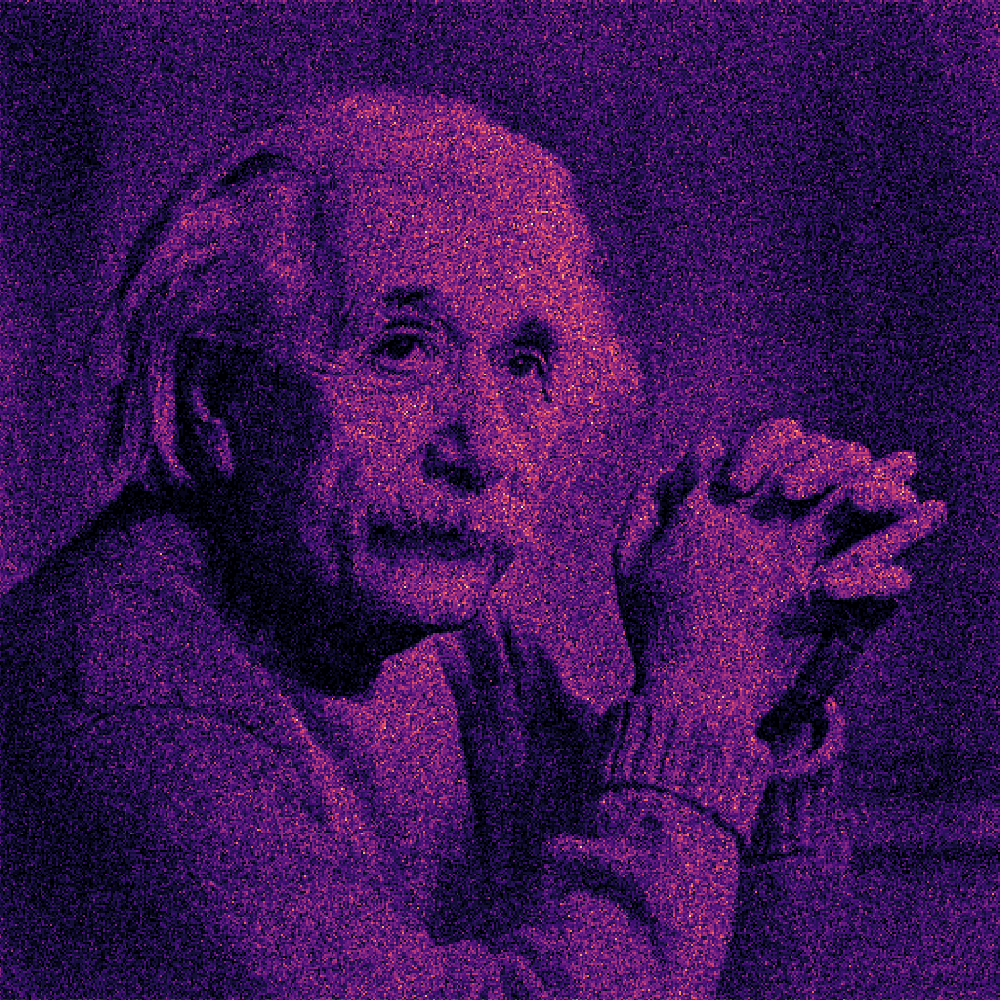}
		\caption*{Data Samples}
		\label{fig:einstein_data}
	\end{subfigure}\hspace*{0.8em}
	\begin{subfigure}{.24\textwidth}
		\centering
		\includegraphics[width=1.0\linewidth]{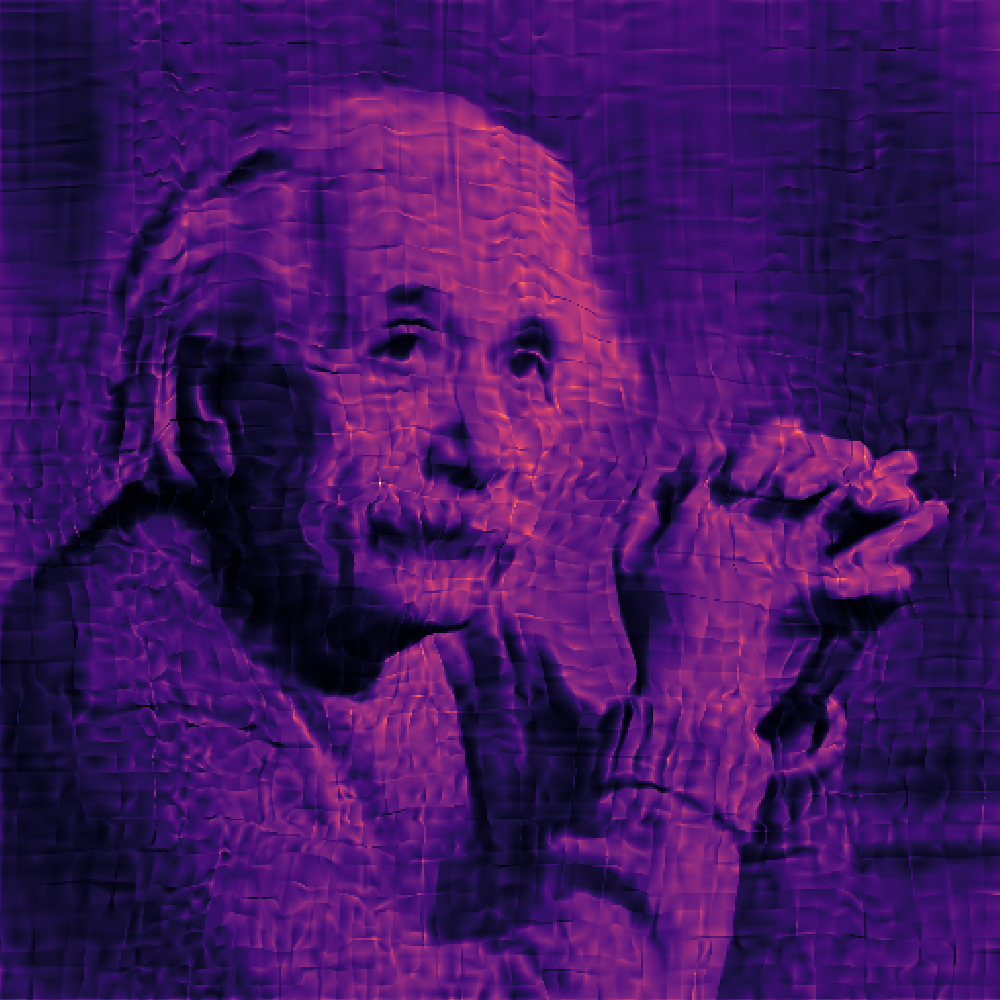}
		\caption*{Estimated Density}
		\label{fig:einstein_density}
	\end{subfigure}\hspace*{0.8em}
	\begin{subfigure}{.24\textwidth}
		\centering
		\includegraphics[width=1.0\linewidth]{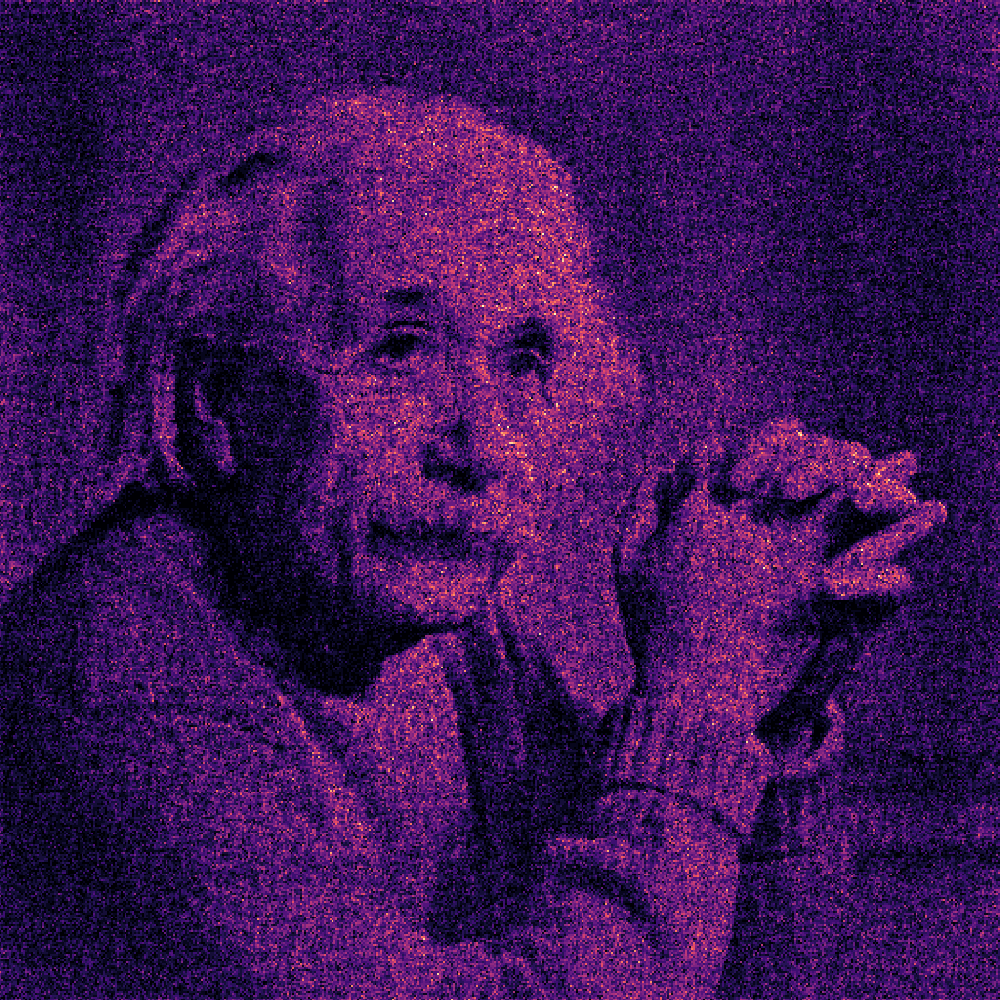}
		\caption*{Model Samples}
		\label{fig:einstein_samples}
	\end{subfigure}
	\caption{Density estimation using data samples of a complicated synthetic distribution derived from a picture. As it can be seen, LRS flows can reconstruct the underlying distribution accurately.}
	\label{fig:einstein}
\end{figure*}

In fact, linear rational splines can perform density estimation tasks on more complicated distributions. Figure \ref{fig:einstein} shows the result of a density estimation task, which involves a highly sophisticated distribution. In both of these experiments, our model consists of two coupling layers constructed by linear rational splines. For detailed information on the configuration used in these simulations, refer to Appendix \ref{ap:density_estimation}.
 
\subsection{Density Estimation of Real-world Data}\label{sec:uci}

For the next experiment, we apply density estimation using maximum likelihood on standard benchmark datasets. Four of these datasets (Power, Gas, HEPMASS, and MiniBooNE) are tabular data from the UCI machine learning repository\footnote{\href{http://archive.ics.uci.edu/ml}{http://archive.ics.uci.edu/ml}}. Furthermore, BSDS300 is a dataset containing patches of natural images~\citep{martin2001adatabase}. We used the preprocessed data of Masked Autoregressive Flows~\citep{papamakarios2017masked}, which is available online\footnote{\href{https://doi.org/10.5281/zenodo.1161203}{https://doi.org/10.5281/zenodo.1161203}}.

Table \ref{tab:uci} shows the test set log-likelihood comparison of our proposed method and FFJORD~\citep{grathwohl2019FFJORD}, Glow~\citep{kingma2018glow}, Masked Autoregressive Flow (MAF)~\citep{papamakarios2017masked}, Neural Autoregressive Flow (NAF)~\citep{huang2018neural}, Block Neural Autoregressive Flow (B-NAF)~\citep{de2019block}, and Neural Spline Flows (NSF). Note that NSF models either use quadratic (Q) piecewise polynomials as in \cite{mueller2019neural} or rational quadratic (RQ) splines as in \cite{durkan2019neural}. Moreover, recall that piecewise polynomials can be used in either coupling or autoregressive modes. These two modes are indicated in this table and the following ones with (C) and (AR), respectively. As suggested in \cite{durkan2019neural}, ResMADE architecture~\citep{durkan2019autoregressive} was used for autoregressive transformations.

It can be seen in Table \ref{tab:uci} that linear rational spline flows perform competitively with RQ-NSF despite using lower degree polynomials.
Furthermore, although here we set the $\lambda^{(k)}$ of each bin to be a single parameter for itself, one could consider a unified parameter to be set for the $\lambda^{(k)}$ of all intervals. While this action can slightly degrade the performance, the results are still comparable to the other methods.

\begin{table*}[tb!]
	\caption{Log-likelihood of the test set in nats (higher is better) for four UCI datasets plus BSDS300. All the results except for our method are taken from the existing literature. In rows with $\dagger$ superscript, the error bar is calculated across independently trained models, while in others it has been calculated on the test set.}
	\label{tab:uci}
	\begin{tabularx}{\textwidth}{@{\extracolsep{\fill}}lccccc}
		\toprule
		\textbf{MODEL}       &\textbf{POWER}               &\textbf{GAS}                 &\textbf{HEPMASS}           &\textbf{MINIBOONE}           &\textbf{BSDS300}\\
 
		\midrule
		FFJORD                                   & 0.46                        & 8.59                        & \SI{-14.92}{}             & \SI{-10.43}{}           & 157.40                \\ 
		Glow                                     & \SI{0.42 \pm 0.01}{}        & \SI{12.24\pm 0.03}{}        & \SI{-16.99 \pm 0.02}{}    & \SI{-10.55\pm 0.45}{}   &\SI{156.95 \pm 0.28}{} \\ 
		Q-NSF (C)                                & \SI{0.64 \pm 0.01}{}        & \SI{12.80\pm 0.02}{}        & \SI{-15.35 \pm 0.02}{}    & \SI{-9.35 \pm 0.44}{}   &\SI{157.65 \pm 0.28}{} \\ 
		RQ-NSF (C)                               & \SI{0.64 \pm 0.01}{}        & \SI{13.09\pm 0.02}{}        & \SI{-14.75 \pm 0.03}{}    & \SI{-9.67 \pm 0.47}{}   &\SI{157.54 \pm 0.28}{} \\
		\midrule 
		Ours (C)                                 & \SI{0.65 \pm 0.01}{}        & \SI{12.99 \pm 0.02}{}       & \SI{-14.64 \pm 0.03}{}    & \SI{-9.65 \pm 0.48}{}   & \SI{157.70 \pm 0.28}{} \\ 
		\midrule
		MAF                                      & \SI{0.45 \pm 0.01}{}        & \SI{12.35\pm 0.02}{}        & \SI{-17.03 \pm 0.02}{}    & \SI{-10.92\pm 0.46}{}   &\SI{156.95 \pm 0.28}{}\\  
		NAF\textsuperscript{$\dagger$}           & \SI{0.62 \pm 0.01}{}        & \SI{11.96\pm 0.33}{}        & \SI{-15.09 \pm 0.40}{}    & \SI{-8.86 \pm 0.15}{}   &\SI{157.73 \pm 0.04}{}\\ 
		B-NAF\textsuperscript{$\dagger$}         & \SI{0.61 \pm 0.01}{}        & \SI{12.06\pm 0.09}{}        & \SI{-14.71 \pm 0.38}{}    & \SI{-8.95 \pm 0.07}{}   &\SI{157.36 \pm 0.03}{}\\
		Q-NSF (AR)                               & \SI{0.66 \pm 0.01}{}        & \SI{12.91\pm 0.02}{}        & \SI{-14.67 \pm 0.03}{}    & \SI{-9.72 \pm 0.47}{}   &\SI{157.42 \pm 0.28}{}\\
		RQ-NSF (AR)                              & \SI{0.66 \pm 0.01}{}        & \SI{13.09\pm 0.02}{}        & \SI{-14.01 \pm 0.03}{}    & \SI{-9.22 \pm 0.48}{}   &\SI{157.31 \pm 0.28}{}\\
		\midrule
		Ours (AR)                                & \SI{0.66 \pm 0.01}{}        & \SI{13.07\pm 0.02}{}        & \SI{-13.80 \pm 0.03}{}    & \SI{-9.77 \pm 0.55}{}   &\SI{158.39 \pm 0.28}{}\\
		\bottomrule
	\end{tabularx}
\end{table*}

\subsection{Generative Modeling of Image Datasets}\label{sec:im_gen}

Next, we perform invertible generative modeling on benchmark image datasets including MNIST~\citep{lecun1998mnist}, CIFAR-10~\citep{krizhevsky2009learning}, ImageNet \textcolor{fe}{$32 \times 32$} and \textcolor{fe}{$64 \times 64$}~\citep{deng2009imagenet, chrabaszcz2017imagenet32}. We measure the performance of our proposed method in bits per dimension, and then compare it with other existing methods. These include Real~NVP~\citep{dinh2016density}, Glow~\citep{kingma2018glow}, \mbox{FFJORD~\citep{grathwohl2019FFJORD}}, i-ResNets~\citep{behrmann2019iresnet}, residual flow~\citep{chen2019residualflows} and rational quadratic spline flows~\citep{durkan2019neural}. The results are given in Table \ref{tab:im_gen}.
\begin{table*}[tb!]
	\centering
	\caption{Bits per dimension (BPD) (lower is better) of the test set for four standard benchmark datasets. All the results except for our method are taken from the existing literature.}
	\label{tab:im_gen}
	\begin{tabularx}{\textwidth}{@{\extracolsep{\fill}}lcccc}
		\toprule
		\textbf{MODEL}                &\textbf{MNIST}               &\textbf{CIFAR-10}                 &\textbf{IMAGENET 32}           &\textbf{IMAGENET 64}    \\ 
		\midrule
		Real~NVP        & \SI{1.06}{}        & \SI{3.49}{}        & \SI{4.28}{}    & \SI{3.98}{}\\ 
		Glow & \SI{1.05}{}        & \SI{3.35}{}        & \SI{4.09}{}    & \SI{3.81}{}\\ 
		FFJORD        & \SI{0.99}{}        & \SI{3.40}{}        & ---    & --- \\ 
		i-ResNet  & \SI{1.05}{}        & \SI{3.45}{}        & ---    & --- \\ 
		Residual Flows       & \SI{0.97}{}        & \SI{3.28}{}        & \SI{4.01}{}    & \SI{3.75}{}\\
		RQ-NSF (C)& ---        & \SI{3.38}{}        & ---    & \SI{3.82}{}\\
		\midrule
		Ours (C)  & 0.91       & 3.38         & 4.09    & 3.82 \\
		\bottomrule
	\end{tabularx}
\end{table*}
\begin{table*}[h]
	\caption{Test set evidence lower bound (ELBO) (higher is better) and negative log-likelihood (NLL) (lower is better) in nats for MNIST~\citep{lecun1998mnist} and EMNIST~\citep{cohen2017emnist}. All the results except for our method are taken from \cite{durkan2019neural}. As in \cite{durkan2019neural}, the NLL is estimated using an importance-weighted approach~\citep{burda2016importance}.}
	\label{tab:vae}
	\begin{tabularx}{\textwidth}{@{\extracolsep{\fill}}lcccc}
		\toprule
		\textbf{}                     &\multicolumn{2}{c}{\textbf{MNIST}}               &\multicolumn{2}{c}{\textbf{EMNIST}}\\
		\cmidrule(lr){2-3}\cmidrule(l){4-5}
		\textbf{MODEL}                &\textbf{ELBO \textuparrow}              &\textbf{NLL \textdownarrow}                 &\textbf{ELBO \textuparrow}           &\textbf{NLL \textdownarrow}\\ 
		\midrule
		Glow       & \SI{-82.25 \pm 0.46}{}        & \SI{79.72 \pm 0.42}{}        & \SI{-120.04 \pm 0.40}{}    & \SI{117.54 \pm 0.38}{}\\ 
		RQ-NSF (C) & \SI{-82.08 \pm 0.46}{}        & \SI{79.63 \pm 0.42}{}        & \SI{-119.74 \pm 0.40}{}    & \SI{117.35 \pm 0.38}{}\\
		\midrule 
		Ours (C)    & \SI{-82.23 \pm 0.46}{}       & \SI{79.66 \pm 0.42}{}        & \SI{-120.46 \pm 0.40}{}    & \SI{117.74 \pm 0.38}{}\\
		\midrule 
		IAF/MAF    & \SI{-82.56 \pm 0.48}{}        & \SI{79.95 \pm 0.43}{}        & \SI{-119.85 \pm 0.40}{}    & \SI{117.47 \pm 0.38}{}\\
		RQ-NSF (AR)& \SI{-82.14 \pm 0.47}{}        & \SI{79.71 \pm 0.43}{}        & \SI{-119.49 \pm 0.40}{}    & \SI{117.28 \pm 0.38}{}\\
		\midrule
		Ours (AR)  & \SI{-82.02 \pm 0.46}{}       & \SI{79.42 \pm 0.42}{}         & \SI{-119.50 \pm 0.39}{}    & \SI{117.23 \pm 0.38}{} \\
		\bottomrule
	\end{tabularx}
\end{table*}
The results in Table \ref{tab:im_gen} demonstrate the competitive performance of linear rational splines with respect to other methods despite their simplicity. Note that Glow uses twice as many parameters as used in neural spline flows including linear rational functions. Furthermore, although residual flows perform better than our proposed method, they require much more computation time in the sampling process where they have to compute the inverse of each invertible ResNet using a fixed point iterative method. In contrast, linear rational splines have an analytic inverse which is also a linear rational spline. This means both the forward and inverse have the same cost. In particular, our experiments indicate that the sampling process in linear rational splines is faster than residual flows by an order of magnitude.

For a detailed information on the experiments and randomly generated sample images of the model, see Appendices \ref{ap:im_gen} and \ref{ap:im_gen_samples}, respectively. \textcolor{fe}{Also, more simulation results can be found in Appendix \ref{ap:extended_simulation_results}.}

\subsection{Variational Auto-encoders}\label{sec:vae}

Finally, we test the performance of our proposed model in a variational auto-encoder (VAE)~\citep{kingma2013vae} setting. In short, in a VAE data points are modeled as realizations of a random variable whose distribution is assumed to be the result of marginalization over a lower-dimensional latent variable.  To make this process tractable, \textcolor{fe}{the type of prior and approximate posterior random variables need to be determined carefully}. Like other flow-based models, linear rational spline flows can be used as effective models for priors and approximate posteriors in a VAE setting. For a detailed explanation on normalizing flows in the context of VAEs, we refer the interested reader to~\citep{rezende2015variational}.

Table \ref{tab:vae} shows the quantitative results of VAE simulation using linear rational splines. As can be seen, our proposed model performs almost as well as other models such as Glow~\citep{kingma2018glow} and rational quadratic splines~\citep{durkan2019neural}. For more details on the experimental configuration and model samples see Appendices \ref{ap:vae} and \ref{ap:vae_samples}, respectively.
 
\section{CONCLUSION}\label{sec:cnclsn}

In this paper, we investigated the use of monotonic linear rational splines in the context of invertible generative modeling. We saw that this family of piecewise functions have the advantage that their inverse is straightforward, and does not require solving degree 2 or 3 polynomial equations. Furthermore, since the same family of functions defines both the forward and inverse, investigation of the mathematical properties of these models is more straightforward. Also, we showed that despite their simplicity, they could perform competitively with more complicated methods in a suite of experiments on synthetic and real datasets.
   
\subsubsection*{Acknowledgements}

\textcolor{fe}{We would like to thank the reviewers for their valuable feedback on our work, helping us to improve the final manuscript.}
 
This research was undertaken using the LIEF HPC-GPGPU Facility hosted at the University of Melbourne. This Facility was established with the assistance of LIEF Grant LE170100200.

\bibliography{references}

\newpage
\appendix
\onecolumn
\thispagestyle{empty}
\aistatstitle{Appendix:\\Invertible Generative Modeling using Linear Rational Splines}
\vspace*{-10em}
\section{MONOTONIC LINEAR RATIONAL SPLINES}

\subsection{Derivative Computation}

Using the quotient rule for derivatives, the derivative of a linear rational spline function (as $g(\phi)$ in Eq. \eqref{eq:lrs_final}) can be computed as:
\begin{equation}\label{eq:ap_lrs_der}
\frac{\mathrm{d}g(\phi)}{\mathrm{d}\phi}=\begin{cases} 
\hfil \dfrac{\lambda^{(k)} w^{(k)} w^{(m)} \big(y^{(m)} - y^{(k)}\big)}{\Big(w^{(k)} \big(\lambda^{(k)}-\phi\big) + w^{(m)} \phi\Big)^2} & 0 \leq \phi \leq \lambda^{(k)} \\ \\
\dfrac{\big(1-\lambda^{(k)}\big) w^{(m)} w^{(k+1)} \big(y^{(k+1)} - y^{(m)}\big)}{\Big(w^{(m)} (1-\phi) + w^{(k+1)} \big(\phi-\lambda^{(k)}\big)\Big)^2} & \lambda^{(k)} \leq \phi \leq 1
\end{cases}
\end{equation}
To calculate the derivative with respect to $x$, we only need to divide Eq. \eqref{eq:ap_lrs_der} by $\delta^{(k)}=x^{(k+1)}-x^{(k)}$. As can be seen, the derivative of the function $g(x)$ never changes sign, even outside the interval $\big[x^{(k)}, x^{(k+1)}\big]$.

\subsection{Inverse Computation}

Unlike rational quadratic splines which require computing the root of a degree two polynomial, linear rational splines have a straightforward closed-form inverse. This function is again a linear rational spline, but with different parameters. The inverse of Eq. \eqref{eq:lrs_final} can be computed as:
\begin{equation}\label{eq:ap_lrs_final_inverse}
g^{-1}(y)=\begin{cases} 
\hfil \dfrac{\lambda^{(k)} w^{(k)} \big(y^{(k)} - y\big)}{w^{(k)} \big(y^{(k)} - y\big) + w^{(m)} \big(y - y^{(m)}\big)} & y^{(k)} \leq y \leq y^{(m)} \\ \\
\dfrac{\lambda^{(k)} w^{(k+1)} \big(y^{(k+1)} - y\big) + w^{(m)} \big(y - y^{(m)}\big)}{w^{(k+1)} \big(y^{(k+1)} - y\big) + w^{(m)} \big(y - y^{(m)}\big)} & y^{(m)} \leq y \leq y^{(k+1)}
\end{cases}
\end{equation}
Again, this function gives us the value of $\phi$ in each interval. We should calculate $x=\delta^{(k)}\phi+x^{(k)}$ to translate this into the interval $\big[x^{(k)}, x^{(k+1)}\big]$.

\subsection{Inverse Derivative Computation}

The derivative of the inverse can be computed using the following relationship:
\begin{equation}\label{eq:ap_lrs_final_inverse_der}
\frac{\mathrm{d}g^{-1}(y)}{\mathrm{d}y}=\begin{cases} 
\hfil \dfrac{\lambda^{(k)} w^{(k)} w^{(m)} \big(y^{(m)} - y^{(k)}\big)}{\Big(w^{(k)} \big(y^{(k)} - y\big) + w^{(m)} \big(y - y^{(m)}\big)\Big)^2} & y^{(k)} \leq y \leq y^{(m)} \\ \\
\dfrac{\big(1-\lambda^{(k)}\big) w^{(m)} w^{(k+1)} \big(y^{(k+1)} - y^{(m)}\big)}{\Big(w^{(k+1)} \big(y^{(k+1)} - y\big) + w^{(m)} \big(y - y^{(m)}\big)\Big)^2} & y^{(m)} \leq y \leq y^{(k+1)}
\end{cases}
\end{equation}
This function captures the change of inverse with respect to $\phi$ in each interval. To translate this into $x$, we should multiply this derivative by $\delta^{(k)}$. As in the forward pass, we can see that the derivative of the inverse does not change its sign even outside the interval $0\leq\phi\leq 1$.

\section{DETAILS OF SIMULATION RESULTS}

\subsection{Synthetic Density Estimation}\label{ap:density_estimation}

For the density estimation task on the Rings dataset in Figure \ref{fig:density_estimation}, we generated a set of 350,000 data points. Then, we used batches of size 512 to train our model, which is a linear rational spline (LRS) flow in the coupling layer mode. The number of coupling layers is 2. For the LRS function of each layer, we used 64 bins and a tail bound of 5. For optimization, we used the Adam~\citep{kingma2015adam} optimizer, with a learning rate of $0.0005$ and cosine annealing~\citep{loshchilov2017sgdr}. Finally, a 2-d standard normal was used as the starting probability distribution.

\begin{figure*}[h!]
\centering
	\begin{subfigure}{.28\textwidth}
		\centering
		\includegraphics[width=1.\linewidth]{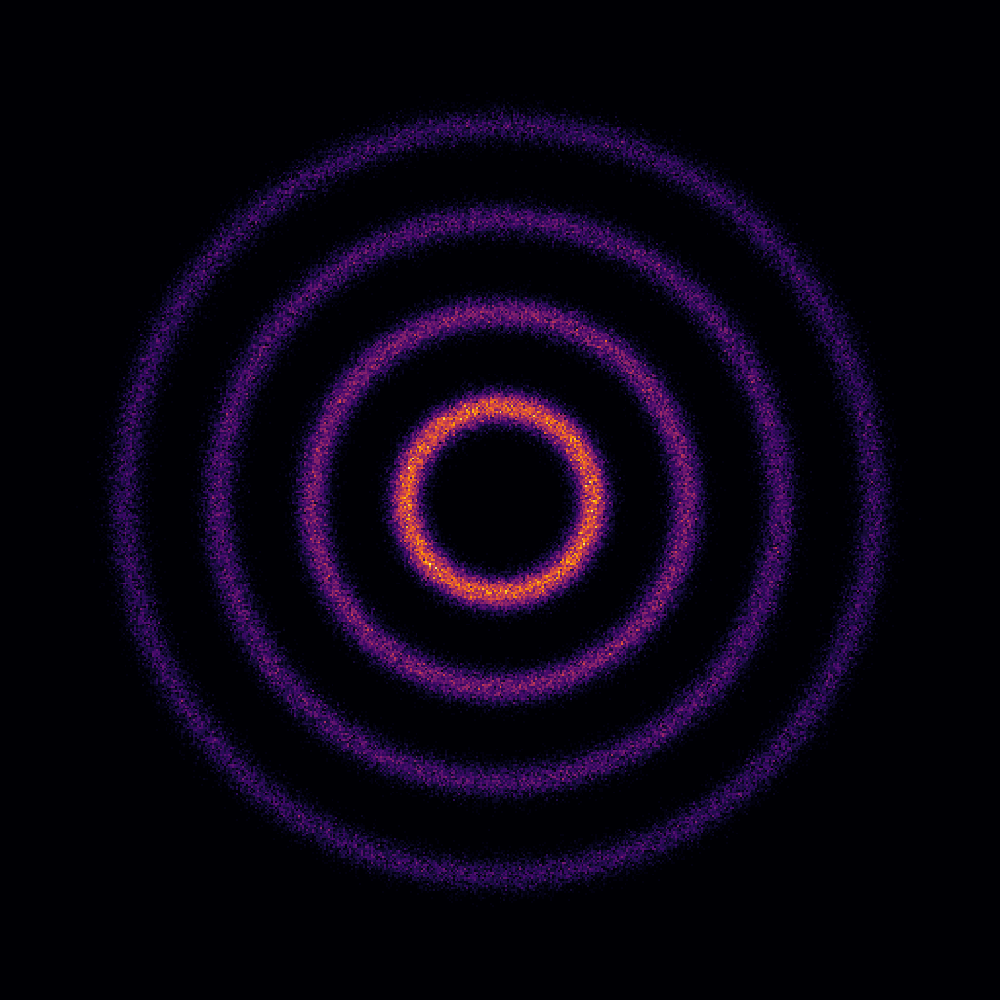}
		\caption*{Data Samples}
		\label{fig:ap_rings_data}
	\end{subfigure}\hspace*{-0.3em}
	\begin{subfigure}{.28\textwidth}
		\centering
		\includegraphics[width=1.\linewidth]{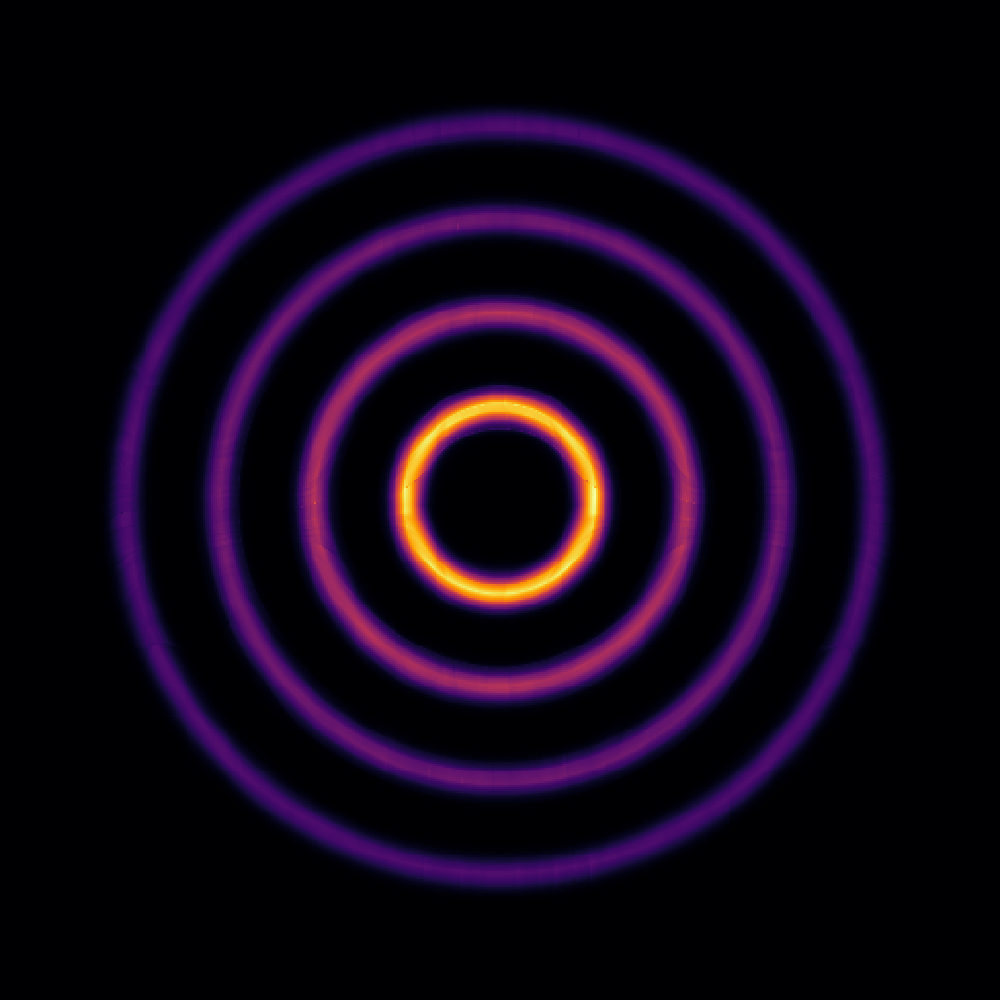}
		\caption*{Estimated Density}
		\label{fig:ap_rings_density}
	\end{subfigure}\hspace*{-0.3em}
	\begin{subfigure}{.28\textwidth}
		\centering
		\includegraphics[width=1.\linewidth]{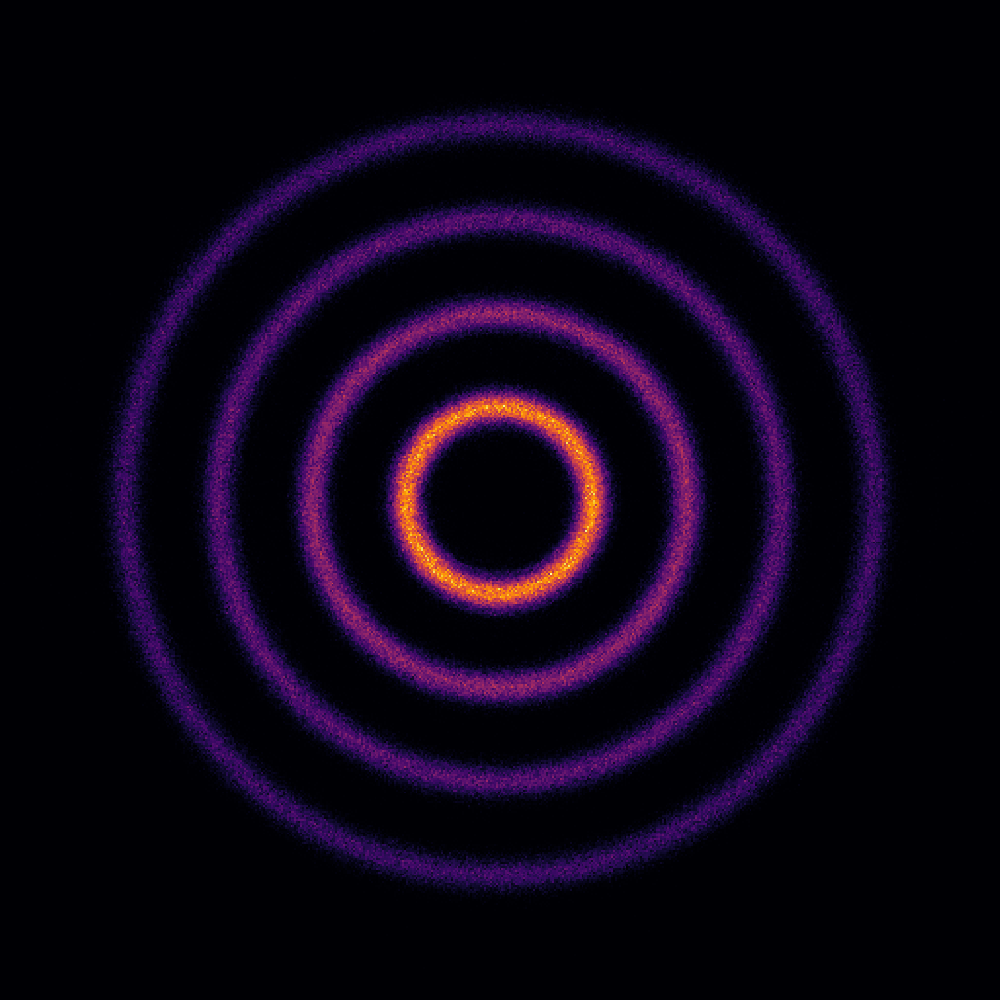}
		\caption*{Model Generated Samples}
		\label{fig:ap_rings_samples}
	\end{subfigure}
	\caption{Density estimation on synthetic 2-d data samples using an infinite data generator.}
	\label{fig:ap_density_estimation}
\end{figure*}

Note that sometimes, it is common to use an infinite data generator, which generates a different set of data at each iteration. We performed our simulation under this condition, too. The results of our method after only 50,000 iterations are depicted in Figure \ref{fig:ap_density_estimation}.

For the results depicted in Figure \ref{fig:einstein}, we used almost the same setup as for the Rings dataset. Here, however, we used a set of $1$M data samples and $1.5$M iterations. Also, we used a uniform random variable as the starting probability distribution.

\subsection{Density Estimation of Real-world Data}\label{ap:uci}

In the density estimation of the UCI tabular datasets and BSDS300~\citep{martin2001adatabase}, we used the configurations of Tables \ref{tab:ap_uci_config_c} and \ref{tab:ap_uci_config_ar} to train our model. Note that here, we used a residual network~\citep{he2016deep} to determine the parameters of a rational linear spline. Like \cite{durkan2019neural}, the Adam optimizer and cosine annealing of the learning rate were used in the optimization of all datasets, except for cases with a $\dagger$ superscript where we used RAdam~\citep{liu2019radam} with no annealing. Unlike~\citep{durkan2019neural}, which uses a fixed value to clip the norm of gradients, we considered changing it here to see if any of the results would be improved. This value has been shown in the tables under \textit{Maximum Gradient Value} row. Also, batch normalization~\citep{ioffe2015batch} and dropout~\citep{srivastava2014dropout} were found to be useful in some of the cases.

\begin{table*}[h!]
	\caption{Hyper-parameters used for simulations of coupling (C) layer transformations using linear rational splines (Table \ref{tab:uci}).}
	\label{tab:ap_uci_config_c}
	\begin{tabularx}{\textwidth}{@{\extracolsep{\fill}}lccccc}
		\toprule
		\textbf{PARAMETER}                &\textbf{POWER}      &\textbf{GAS}       &\textbf{HEPMASS}      &\textbf{MINIBOONE}     &\textbf{BSDS300}\textsuperscript{$\dagger$}\\
		\midrule
		Learning Rate                     &\SI{0.0005}         &\SI{0.0005}        &\SI{0.0005}           &\SI{0.0001}            &\SI{0.0005}{}\\
		Batch Size                        &\SI{512}            &\SI{512}           &\SI{256}              &\SI{128}               &\SI{512}{}\\
		Number of Learning Iterations     &\SI{400}{}k         &\SI{400}{}k        &\SI{400}{}k           &\SI{200}{}k            &\SI{500}{}k\\
		\midrule
		Transformation Layers             &\SI{10}             &\SI{10}            &\SI{20}               &\SI{10}                &\SI{20}{}\\
		Tail Bound                        &\SI{3}              &\SI{3}             &\SI{3}                &\SI{8}                 &\SI{3}{}\\
		Number of Bins                    &\SI{8}              &\SI{8}             &\SI{8}                &\SI{4}                 &\SI{8}{}\\
		ResNet Layers                     &\SI{2}              &\SI{2}             &\SI{1}                &\SI{1}                 &\SI{1}{}\\
		ResNet Hidden Features            &\SI{256}            &\SI{256}           &\SI{128}              &\SI{32}                &\SI{128}{}\\
		\midrule
		Maximum Gradient Value            &\SI{5}              &\SI{5}             &\SI{5}                &\SI{25}                &\SI{5}{}\\
		Dropout Probability               &\SI{0.0}            &\SI{0.1}           &\SI{0.2}              &\SI{0.5}               &\SI{0.5}{}\\
		Batch Normalization               &N                   &N                  &Y                     &Y                      &Y\\		

		\bottomrule
	\end{tabularx}
\end{table*}

\begin{table*}[h!]
	\caption{Hyper-parameters used for simulations of autoregressive (AR) transformations using linear rational splines (Table \ref{tab:uci}).}
	\label{tab:ap_uci_config_ar}
	\begin{tabularx}{\textwidth}{@{\extracolsep{\fill}}lccccc}
		\toprule
		\textbf{PARAMETER}                &\textbf{POWER}      &\textbf{GAS}       &\textbf{HEPMASS}\textsuperscript{$\dagger$}      &\textbf{MINIBOONE}\textsuperscript{$\dagger$}     &\textbf{BSDS300}\textsuperscript{$\dagger$}\\
		\midrule
		Learning Rate                     &\SI{0.0005}         &\SI{0.0005}        &\SI{0.0005}           &\SI{0.0001}            &\SI{0.0005}{}\\
		Batch Size                        &\SI{512}            &\SI{512}           &\SI{512}              &\SI{128}               &\SI{512}{}\\
		Number of Learning Iterations     &\SI{400}{}k         &\SI{400}{}k        &\SI{500}{}k           &\SI{150}{}k            &\SI{500}{}k\\
		\midrule
		Transformation Layers             &\SI{10}             &\SI{10}            &\SI{10}               &\SI{10}                &\SI{10}{}\\
		Tail Bound                        &\SI{3}              &\SI{3}             &\SI{5}                &\SI{5}                 &\SI{3}{}\\
		Number of Bins                    &\SI{8}              &\SI{8}             &\SI{8}                &\SI{8}                 &\SI{8}{}\\
		ResNet Layers                     &\SI{2}              &\SI{2}             &\SI{1}                &\SI{2}                 &\SI{2}{}\\
		ResNet Hidden Features            &\SI{256}            &\SI{256}           &\SI{128}              &\SI{64}                &\SI{512}{}\\
		\midrule
		Maximum Gradient Value            &\SI{5}              &\SI{5}             &\SI{20}               &\SI{20}                &\SI{5}{}\\
		Dropout Probability               &\SI{0.0}            &\SI{0.1}           &\SI{0.5}              &\SI{0.5}               &\SI{0.5}{}\\
		Batch Normalization               &N                   &Y                  &N                     &Y                      &Y\\		
		
		\bottomrule
	\end{tabularx}
\end{table*}

\textcolor{edited}{Also, in Table \ref{tab:ap_uci_nsf} we have included the results of our model under the hyper-parameters set for rational quadratic spline flows~\citep{durkan2019neural}.}

\begin{table*}[h!]
	\caption{\textcolor{edited}{Test set log-likelihood in nats (higher is better) for four UCI datasets plus BSDS300~\citep{martin2001adatabase} under parameters set for rational quadratic splines in~\citep{durkan2019neural}}}
	\label{tab:ap_uci_nsf}
	\begin{tabularx}{\textwidth}{@{\extracolsep{\fill}}lccccc}
		\toprule
		\textbf{MODEL}       &\textbf{POWER}               &\textbf{GAS}                 &\textbf{HEPMASS}           &\textbf{MINIBOONE}           &\textbf{BSDS300}\\
		
		\midrule
		LRS Flows (C)         & \SI{0.65 \pm 0.01}{}        & \SI{12.99 \pm 0.02}{}       & \SI{-14.88 \pm 0.03}{}    & \SI{-9.91 \pm 0.53}{}   & \SI{157.56 \pm 0.28}{} \\
		\midrule
		LRS Flows (AR)        & \SI{0.66 \pm 0.01}{}        & \SI{13.05\pm 0.02}{}        & \SI{-14.37 \pm 0.03}{}    & \SI{-10.63 \pm 0.47}{}   &\SI{157.50 \pm 0.28}{}\\
		\bottomrule
	\end{tabularx}
\end{table*}

\subsection{Generative Modeling of Image Datasets}\label{ap:im_gen}

For the generative modeling tasks, we used the Adam optimizer with cosine annealing of the learning rate. The initial learning rate was set to $0.0005$. For all datasets, we used batches of size $256$, and trained the model for $200$k iterations. We followed the multi-scale architecture of \cite{dinh2016density} as used in rational quadratic splines~\citep{durkan2019neural} and Glow~\citep{kingma2018glow}. As in Glow, each layer consists of multiple stacked steps of basic transformations, which are built by using an actnorm, a 1x1 convolution, and a coupling layer. Here, we used rational linear spline functions to build the coupling layer transformation of each layer. Moreover, a ResNet with batch normalization was used to determine the parameters of each layer's linear rational spline functions. The detailed configuration used for the simulation of each dataset is given in Table \ref{tab:ap_im_gen_config}.

\begin{table*}[h!]
	
	\caption{Hyper-parameters used for invertible generative modeling simulations of Section \ref{sec:im_gen} (Table \ref{tab:im_gen}).}
	\label{tab:ap_im_gen_config}
	\begin{tabularx}{\textwidth}{@{\extracolsep{\fill}}lcccc}
		\toprule
		\textbf{PARAMETER}                &\textbf{MNIST}      &\textbf{CIFAR-10}       &\textbf{IMAGENET 32}        &\textbf{IMAGENET 64}\\
		\midrule
		Learning Rate                     &\SI{0.0005}         &\SI{0.0005}             &\SI{0.0005}                 &\SI{0.0005}{}\\
		Batch Size                        &\SI{256}            &\SI{256}                &\SI{256}                    &\SI{256}{}\\
		Number of Learning Iterations     &\SI{200}{}k         &\SI{200}{}k             &\SI{200}{}k                 &\SI{200}{}k\\
	    Validation Frac. of Train. Set    &\SI{2}{}\%          &\SI{1}{}\%              &\SI{2}{}\%                  &\SI{1}{}\%\\
		\midrule
		Multi-scale Transform Levels      &\SI{2}              &\SI{3}                  &\SI{4}                      &\SI{4}{}\\
        Num. of Trans. per Layer          &\SI{7}              &\SI{7}                  &\SI{7}                      &\SI{7}{}\\
 		ResNet Blocks                     &\SI{3}              &\SI{3}                  &\SI{3}                      &\SI{3}{}\\
        ResNet Hidden Channels            &\SI{128}            &\SI{96}                 &\SI{128}                    &\SI{96}{}\\
		Tail Bound                        &\SI{3}              &\SI{3}                  &\SI{3}                      &\SI{3}{}\\
		Number of Bins                    &\SI{32}             &\SI{4}                  &\SI{32}                     &\SI{8}{}\\
		\midrule
		Dropout Probability               &\SI{0.2}            &\SI{0.1}                &\SI{0.2}                    &\SI{0.0}{}\\	
		
		\bottomrule
	\end{tabularx}
\end{table*}

\subsection{Variational Auto-Encoders}\label{ap:vae}

For variational auto-encoders, we follow the same procedure as neural spline flows~\citep{durkan2019neural}. First, a linear warm-up multiplier is used for the KL-divergence term in the cost function. This multiplier starts at the value $0.5$, and then linearly increases to $1$ as $10$\% of the training set passes. A ResNet with $2$ blocks determines the parameters of the linear rational splines used in either coupling (C) or autoregressive (AR) transformations. The dimension of the latent space is set to $32$, and $64$ context features are computed by the encoder.  

As before, the Adam optimizer with cosine annealing of an initial $0.0005$ learning rate is used for optimization. We use batches of size $256$, and train the model for $150$k iterations. Model selection is made using a validation set of $10$k and $20$k samples for MNIST and EMNIST, respectively.

\clearpage
\section{IMAGE SAMPLES}

\subsection{Random Image Samples Generated by Models Trained for Section \ref{sec:im_gen}}\label{ap:im_gen_samples}

\begin{center}
	\begin{figure*}[h]
		\centering
		\begin{subfigure}{.40\textwidth}
			\centering
			\includegraphics[width=1.0\linewidth]{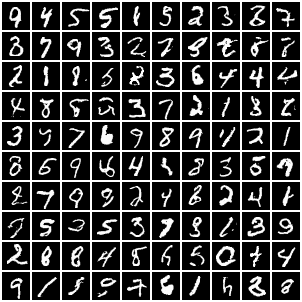}
			\caption*{MNIST}
			\label{fig:ap_mnist}
		\end{subfigure}\hspace*{1.5em}
		\begin{subfigure}{.40\textwidth}
			\centering
			\includegraphics[width=1.0\linewidth]{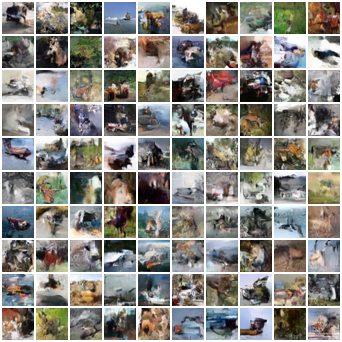}
			\caption*{CIFAR-10}
			\label{fig:ap_cifar_10}
		\end{subfigure}\\\vspace*{1em}
		\begin{subfigure}{.40\textwidth}
			\centering
			\includegraphics[width=1.0\linewidth]{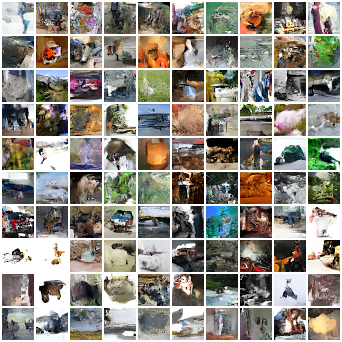}
			\caption*{ImageNet 32x32}
			\label{fig:ap_imagenet32}
		\end{subfigure}\hspace*{1.5em}
		\begin{subfigure}{.40\textwidth}
			\centering
			\includegraphics[width=1.0\linewidth]{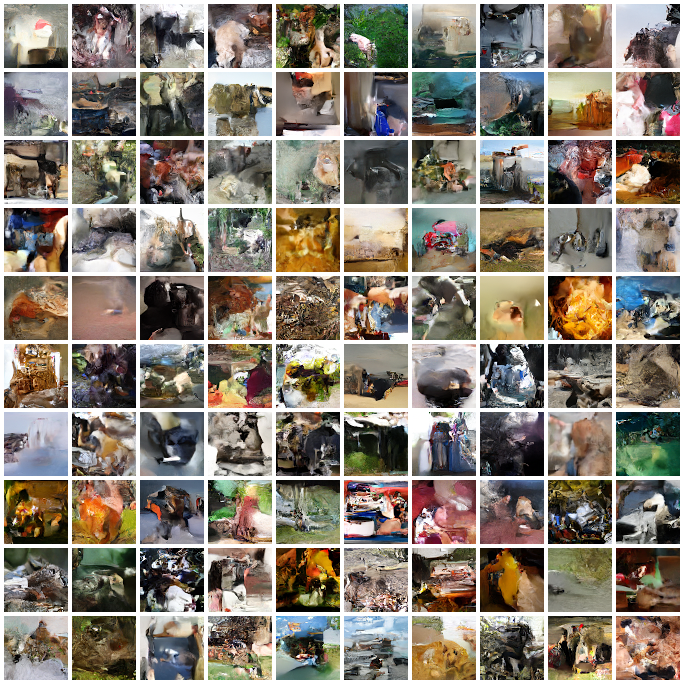}
			\caption*{ImageNet 64x64}
			\label{fig:ap_imagenet64}
		\end{subfigure}
		\caption{Random image samples generated by trained linear rational spline flows.}
		\label{fig:ap_im_gen_samples}
	\end{figure*}
\end{center}

\newpage
\subsection{Randomly Generated VAE Samples}\label{ap:vae_samples}
\vspace{2em}
\begin{center}
	\begin{figure*}[h]
		\centering
		\begin{subfigure}{.4\textwidth}
			\centering
			\includegraphics[width=1.0\linewidth]{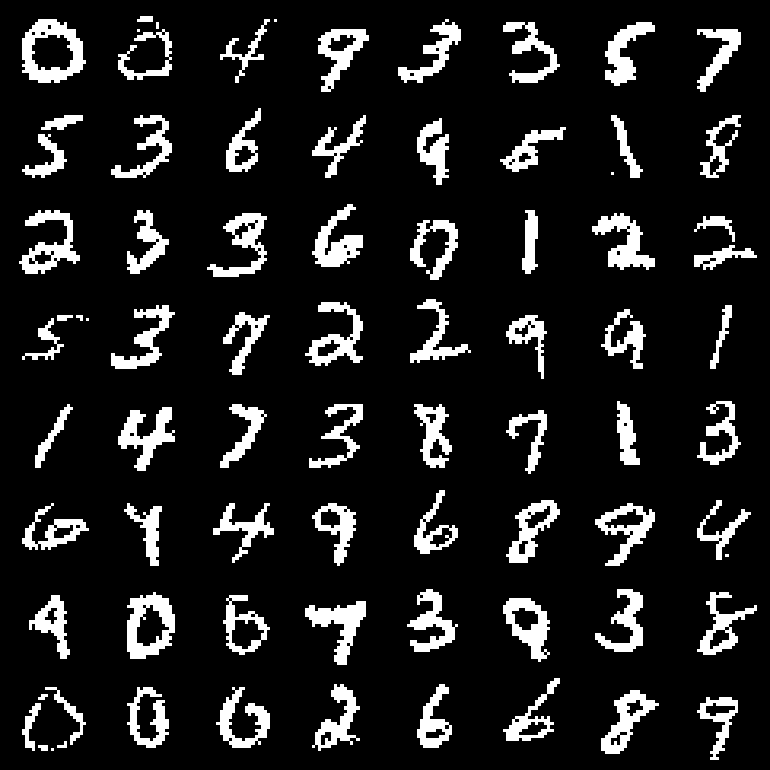}
			\caption*{MNIST}
			\label{fig:ap_vae_mnist}
		\end{subfigure}\\\vspace*{1em}
		\begin{subfigure}{.4\textwidth}
			\centering
			\includegraphics[width=1.0\linewidth]{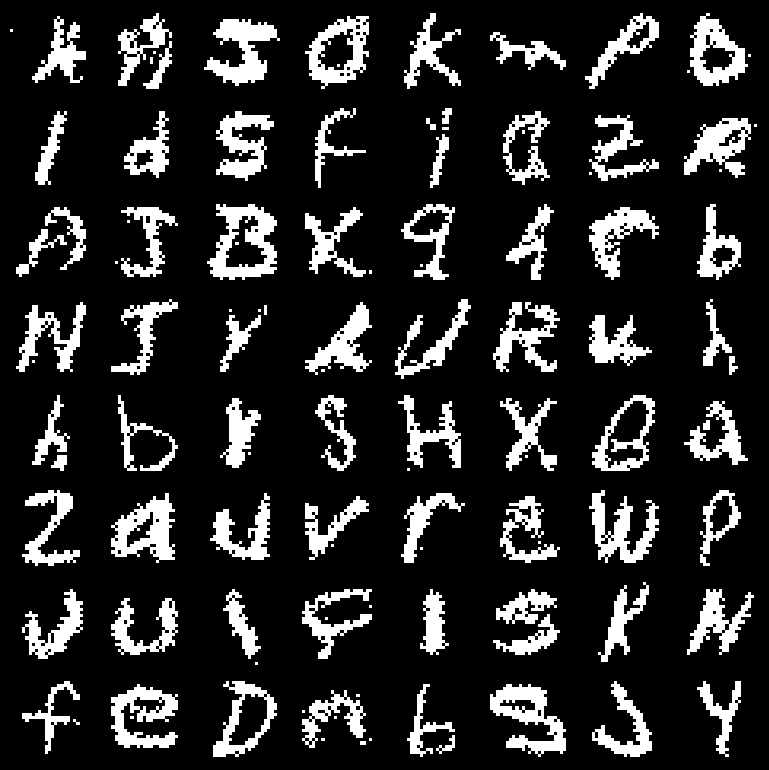}
			\caption*{EMNIST}
			\label{fig:ap_vae_emnist}
		\end{subfigure}
		\caption{Randomly generated VAE image samples. Linear rational spline flows were used as the prior and approximate posterior in these models.}
	    \label{fig:ap_vae_samples}
	\end{figure*}
\end{center}

\newpage
\section{Further Simulation Results}\label{ap:extended_simulation_results}

\textcolor{fe}{To highlight the improvements in the current work, we perform a new set of image generation experiments on the MNIST~\citep{lecun1998mnist} dataset. Other than using a different family of splines, all of the hyperparameters of the models (summarized in Table \ref{tab:ap_hyper_extended}) are fixed to be the same. For a given depth, the experiment is performed for 8 different seeds. We then train the model and repeat the same procedure for 5 various depths. In each of the experiments we pick the best flow model using a validation set. Finally, we measure the log-likelihood on the test set in BPD.}

\begin{table*}[h!]
	\centering
	\caption{Hyper-parameters used for invertible generative modeling simulations of Fig. \ref{fig:ap_img_gen_mnist}.}
	\label{tab:ap_hyper_extended}
	\begin{tabularx}{0.45\textwidth}{lc}
		\toprule
		\textbf{PARAMETER}                &\textbf{MNIST}\\
		\midrule
		Learning Rate                     &\SI{0.001}{}\\
		Batch Size                        &\SI{256}{}\\
		Number of Learning Iterations     &\SI{50}{}k\\
		Validation Frac. of Train. Set    &\SI{2}{}\%\\
		\midrule
		Multi-scale Transform Levels      &\SI{1}{}\\
		Num. of Trans. per Layer          &$2, 4, 8, 16, 32$\\
		ResNet Blocks                     &\SI{2}{}\\
		ResNet Hidden Channels            &\SI{96}{}\\
		Tail Bound                        &\SI{3}{}\\
		Number of Bins                    &\SI{4}{}\\
		\midrule
		Dropout Probability               &\SI{0.2}{}\\	
		
		\bottomrule
	\end{tabularx}
\end{table*}

\begin{center}
	\begin{figure*}[h]
		\centering
		\begin{subfigure}{.5\textwidth}
			\centering
			\includegraphics[width=1.0\linewidth]{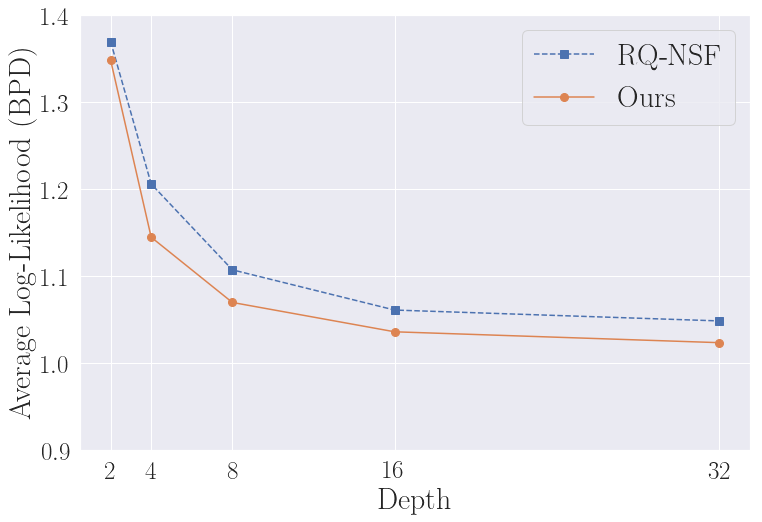}
			\label{fig:ap_mnist_bpd}
		\end{subfigure}\hspace*{1.5em}
		\begin{subfigure}{.5\textwidth}
			\centering
			\includegraphics[width=1.0\linewidth]{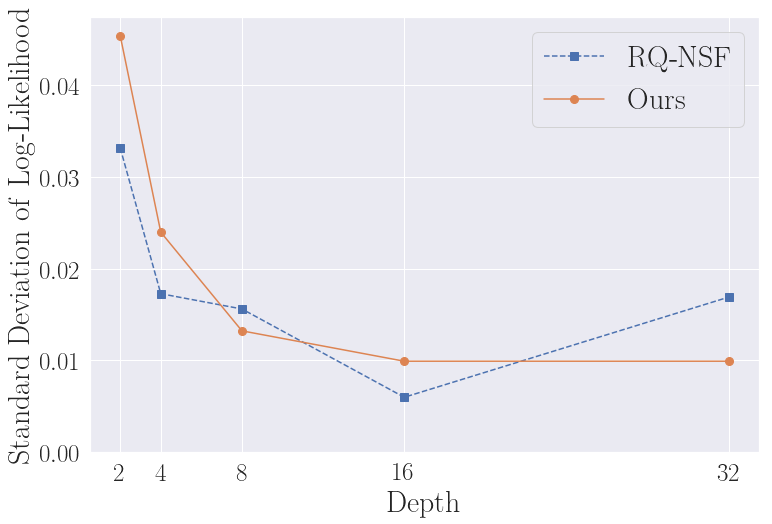}
			\label{fig:ap_mnist_bpd_var}
		\end{subfigure}\\\vspace*{1em}
		\begin{subfigure}{.5\textwidth}
			\centering
			\includegraphics[width=1.0\linewidth]{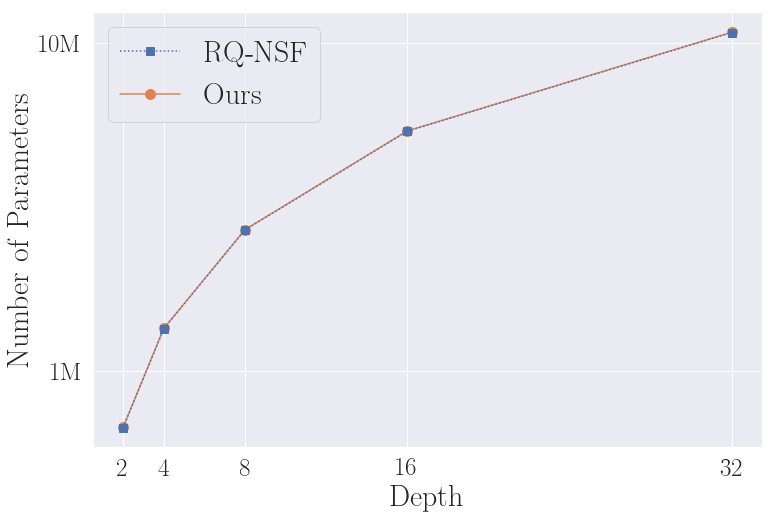}
			\label{fig:ap_mnist_param_num}
		\end{subfigure}\hspace*{1.5em}
		\begin{subfigure}{.5\textwidth}
			\centering
			\includegraphics[width=1.0\linewidth]{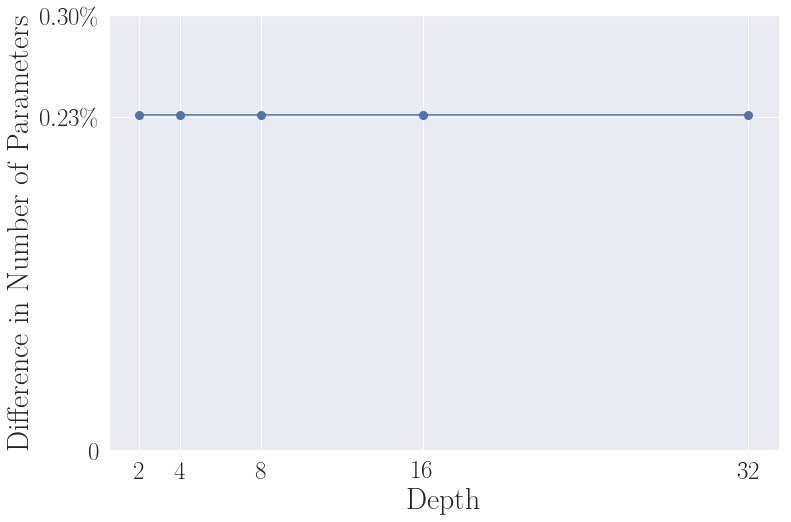}
			\label{fig:ap_mnist_param_dif}
		\end{subfigure}
		\caption{Comparison of linear rational and rational quadratic splines for image generation task on MNIST.}
		\label{fig:ap_img_gen_mnist}
	\end{figure*}
\end{center}

\textcolor{fe}{Figure \ref{fig:ap_img_gen_mnist} shows the simulation results. The top-left figure shows the average of log-likelihood on the 8 seeds. As shown, linear rational splines consistently perform better than rational quadratic splines despite using lower degree polynomials. The top-right figure shows the standard deviation of the results across different seeds. As the figure shows, the standard deviation of linear rational splines consistently decreases as the depth increases. However, the results of rational quadratic splines show fluctuations, and for the depth of 32 their standard deviation gets worse. This might be an indication of the fact that since they are using higher degree polynomials, they require more numerical accuracy as the depth increases. In contrast, as a composition of linear rational splines is still a linear rational spline, the standard deviation of our method's results consistently decreases. Finally, you can see the number of parameters, and its relative change in percentages for these simulations in the bottom figures. As the figures show, the increase in number of parameters is only 0.23\% for this set of simulations which is negligible.}

\end{document}